\documentclass[10pt,twocolumn,letterpaper]{article}

\usepackage[pagenumbers]{cvpr} %

\usepackage{graphicx}
\usepackage{cuted}
\usepackage{caption}

\definecolor{cvprblue}{rgb}{0.21,0.49,0.74}
\usepackage[pagebackref,breaklinks,colorlinks,allcolors=cvprblue]{hyperref}

\def\confName{CVPR}
\def\confYear{2027}

\title{PanDWM: Panoramic 4D Driving World Models}

\author{
Jiajun Sun$^{1*}$ \quad
Zeyu Zhang$^{2*\dag}$ \quad
Zeyu Ren$^{3}$ \quad
Guangfeng Jiang$^{2}$ \quad
Yang Zhao$^{4}$ \quad
Hao Tang$^{2\ddag}$ \\
$^{1}$Shenzhen University \quad
$^{2}$Peking University \quad
$^{3}$The University of Melbourne \quad
$^{4}$La Trobe University \\
\small $^*$Equal contribution.
$^\dag$Project lead.
$^\ddag$Corresponding author: bjdxtanghao@gmail.com.
}

\begin{document}
\renewcommand{\paragraph}[1]{\vspace{.5em}\noindent\textbf{#1.}}
\maketitle

\begin{strip}
  \centering
  \includegraphics[width=0.96\textwidth]{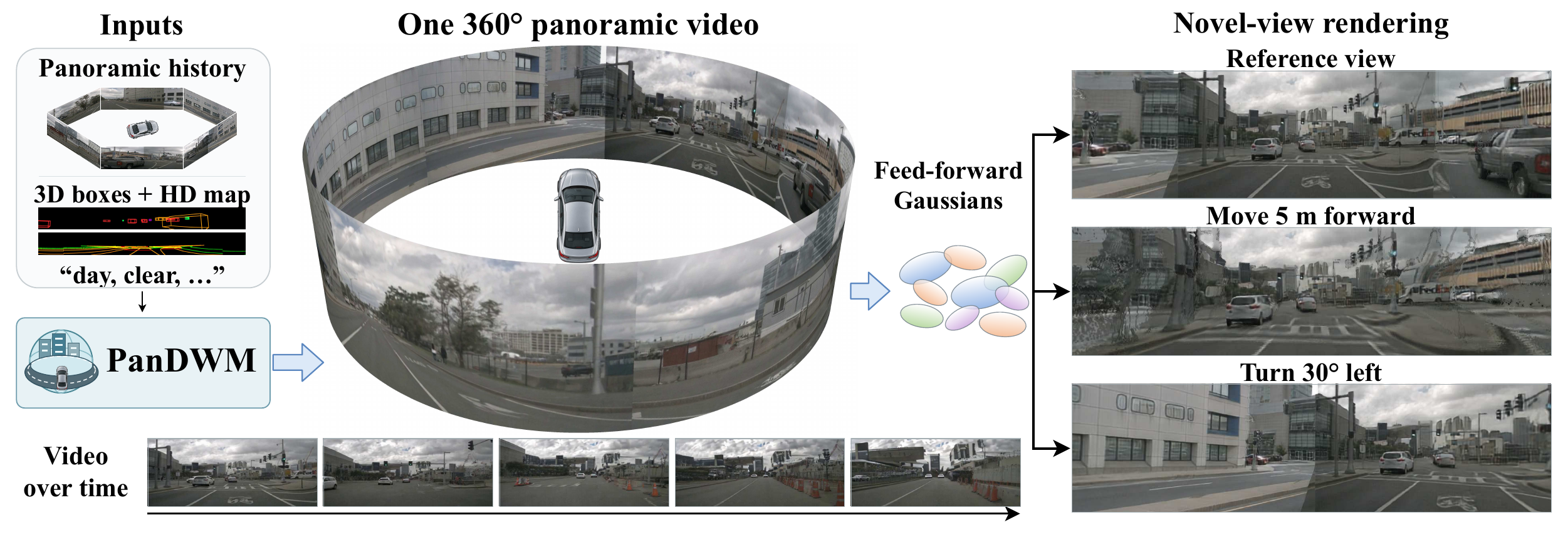}
  \captionof{figure}{PanDWM generates driving video as one 360$^\circ$ panorama and lifts it into Gaussians for novel-view rendering.Given panoramic history, 3D layouts and text, it predicts future panoramas; the ring shows one generated panorama and the bottom row its forward region over time.The right column renders the Gaussian scene of the same frame from the reference viewpoint, 5\,m forward and turned $30^\circ$ left.}
  \label{fig:teaser}
  \vspace{1em}
\end{strip}

\begin{abstract}
Driving world models usually generate videos from six pinhole cameras with separate optical centres and small overlaps, and must learn to keep these views consistent.
We present PanDWM, which stitches them into one 360$^\circ$ panoramic band.
Camera boundaries then fall inside one image, so keeping the views consistent means keeping the image continuous, and each pixel still records its camera ray.
Conditioned on layouts, a causal student generates the band window by window in four steps; it is distilled from a causal teacher and trained on its own rollouts.
A feed-forward decoder lifts each frame along these rays into 3D Gaussians, and static Gaussians from neighbouring frames complete them into 4D scenes.
On nuScenes, scored in one stitched 360$^\circ$ view, PanDWM has lower FID and FVD than the multi-camera baselines at 8--46$\times$ their speed, and keeps its 50-step teacher's FID while lowering FVD by 22--25\%.
Code:~\url{https://github.com/AIGeeksGroup/PanDWM}.
Website:~\url{https://aigeeksgroup.github.io/PanDWM}.
\end{abstract}

\section{Introduction}
\label{sec:intro}

Driving world models predict future surround-view video from scene layouts and past frames, and serve as controllable simulators for autonomous driving.
Their predictions must stay consistent across cameras and over long rollouts.
Simulating trajectories that leave the recorded one also requires rendering the predicted scene from new viewpoints.

Existing layout-conditioned generators~\cite{magicdrivev2,panacea,drivedreamer2,unimlvg} represent the surround view as six perspective videos, one per camera.
This representation cuts the scene at every camera boundary.
The cameras have distinct optical centres and overlap only marginally, and content that crosses a boundary is split between two separately generated images.
Cross-view consistency then has to be enforced by attention between views, by conditioning some views on others, or by concatenating the views into one canvas.
Methods that lift generated video into 3D or 4D scenes~\cite{worldsplat,dist4d,xscene} reconstruct from these per-camera videos and inherit their seams.

We propose PanDWM, a driving world model that generates the surround view as a single 360$^\circ$ panorama (\cref{fig:teaser}).
Instead of aligning six separately generated views, PanDWM places every camera boundary inside one image, where cross-camera consistency becomes spatial continuity.
The stitching map also gives each pixel a source camera and its calibrated ray, which later anchor the 3D Gaussians.
The network processes the panorama as two halves rather than six camera views, which also lowers the cost.
Because camera images sample their periphery more densely, six views that are nowhere coarser than the panorama need 6240 tokens per frame against 4608, and the same network denoises and decodes the halves 1.17$\times$ faster (supplementary \cref{sec:suppl:matchres}).
Long videos are usually produced clip by clip~\cite{dreamforge,glad,unimlvg}, at tens of denoiser evaluations per clip and with a gap between the recorded history seen in training and the generated history seen at inference~\cite{selfforcing}.
PanDWM instead generates the panorama causally, one window at a time, with a four-step student trained on self-generated history.
A feed-forward decoder lifts each generated frame along the calibrated rays into 3D Gaussians for novel-view rendering.
4D Gaussians add ego-aligned static Gaussians from neighbouring frames, filling regions that the current frame does not observe along new trajectories.

Our contributions are:
\begin{itemize}
  \item \textbf{Panoramic driving world modeling.}
  We generate the surround view as one 360$^\circ$ panorama, which turns cross-camera consistency into spatial continuity at a lower cost than six views.
  In the 360$^\circ$ view, PanDWM has lower FID and FVD than multi-camera baselines.
  \item \textbf{A four-step causal panoramic generator.}
  A block-causal student trained on its own rollouts runs 8--46$\times$ faster than the baselines and matches the FID of its 50-step teacher with 4 instead of 100 denoiser evaluations.
  \item \textbf{Feed-forward 3D and 4D Gaussian lifting.}
  A decoder lifts generated latents along the source-camera rays into pixel-aligned Gaussians in one pass, and static Gaussians from neighbouring frames complete them into 4D scenes with the lowest depth error among camera-only methods.
\end{itemize}

\section{Related work}
\label{sec:related}

\paragraph{Driving video generation and world models}
MagicDrive~\cite{magicdrive} introduced layout control for street-view images, conditioning on 3D boxes, road maps and text.
Multi-camera video generators carry these conditions to six views~\cite{magicdrivev2,panacea,drivewm,unimlvg,opendwm,drivescape} and keep the views consistent through factorized cross-camera attention~\cite{panacea} or by placing the perspective images side by side on one canvas~\cite{drivedreamer2}.
Others also predict semantics and metric depth~\cite{omninwm} or bring occupancy and LiDAR into the generation~\cite{uniscene,genesis}.
Longer videos come from conditioning on past frames~\cite{dreamforge,glad,vista,epona,gem}, and recent models distil the generator to a few denoising steps to lower the cost of such rollouts~\cite{horizondrive,m4world,helloworld}.
The panoramas of Panacea and OmniNWM are sets of surround cameras of this kind.
In contrast, PanDWM generates the rig as one panorama in which every camera boundary lies inside the image.

\paragraph{Panoramic generation}
Panoramic video models build on perspective video diffusion.
They adapt pretrained models to equirectangular distortion and the wrap-around seam~\cite{dvd360,panowan}, outpaint perspective clips to the full sphere~\cite{imagine360,argus}, or generate perspective views jointly before stitching them~\cite{videopanda}.
DynamicScaler~\cite{dynamicscaler} slides its denoising window over canvases of arbitrary size, and PanoWorld~\cite{panoworld} follows camera trajectories in aerial scenes.
Each frame in these models has a single projection centre, and no camera rig or driving layout is involved.
In driving, panoramas have served BEV mapping~\cite{onebev,dur360bev} and controllable street-view images~\cite{percep360}, and QuaDreamer~\cite{quadreamer} generates panoramic video for quadruped robots.
None of them generates panoramic driving video from scene layouts.

\paragraph{Few-step autoregressive video diffusion}
Autoregressive video diffusion assigns noise levels per token~\cite{diffusionforcing} or lets noise grow across temporal chunks for streaming~\cite{magi1}.
Few-step causal generators are distilled from a slower teacher, by distribution matching distillation (DMD)~\cite{dmd,dmd2} from a bidirectional teacher in CausVid~\cite{causvid} or by consistency distillation~\cite{cm} from an autoregressive teacher in Causal Forcing++~\cite{causalforcingpp}.
A student trained only on recorded context drifts once it conditions on its own outputs.
Self Forcing~\cite{selfforcing} therefore rolls the student out during training, CausalDrive~\cite{causaldrive} combines this with DMD for real-time front-view driving, and Self-Forcing++~\cite{selfforcingpp} and LongLive~\cite{longlive} extend generation to minutes.
Adversarial losses further sharpen few-step samples~\cite{add,dmd2,apt}.
PanDWM brings self-rollout distillation with an adversarial loss on rollouts to layout-conditioned panoramic driving video.

\begin{figure*}[t]
  \centering
  \includegraphics[width=\textwidth]{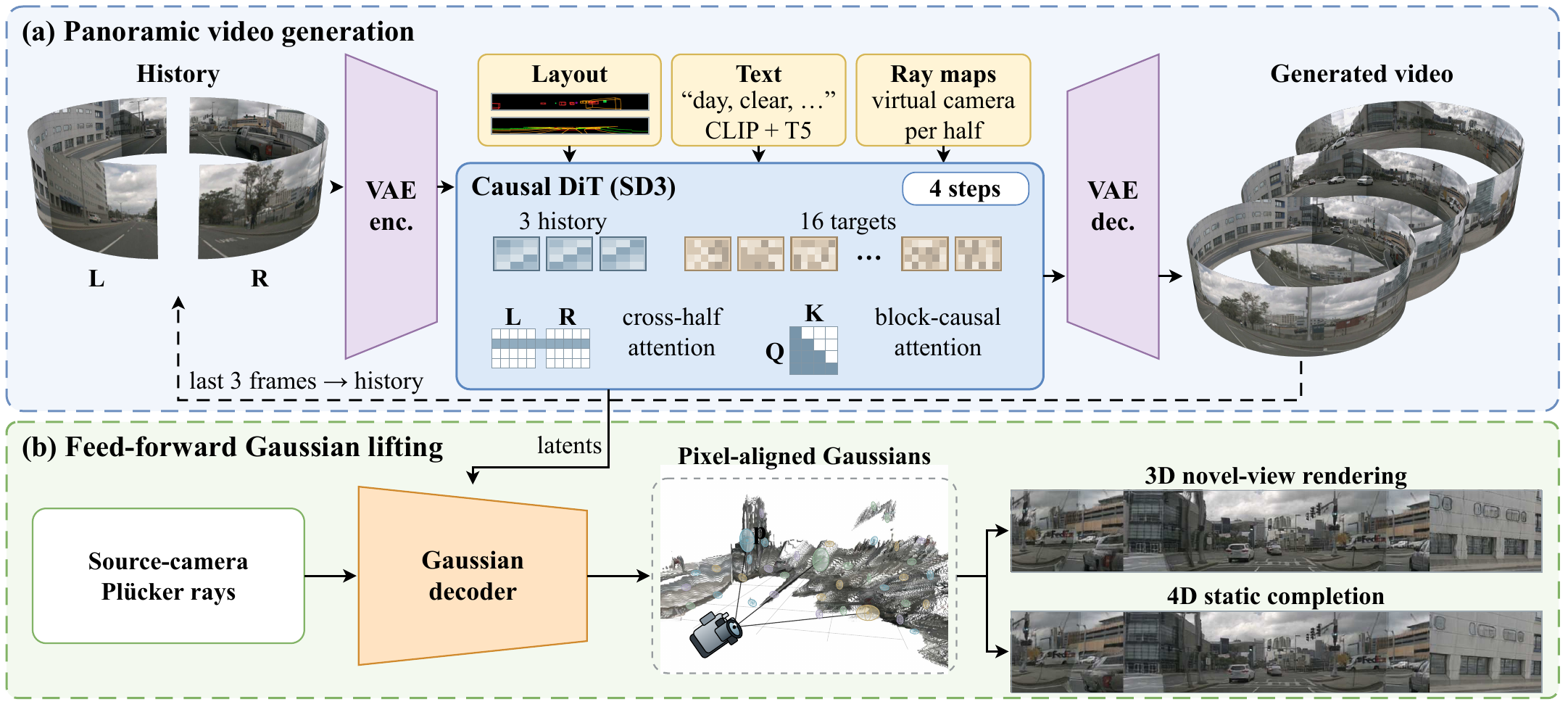}
  \caption{PanDWM overview.
  (a) Panoramic history is split into overlapping halves and encoded by a VAE. Conditioned on layouts, text and ray maps, a causal diffusion transformer predicts 16 frames from 3 history frames in four steps; its final 3 frames seed the next window.
  (b) A feed-forward decoder lifts frame latents into pixel-aligned Gaussians along calibrated source-camera rays, at ranges relabelled with Depth Anything 3. Static geometry from neighbouring frames supports 4D completion.
  Panoramas are displayed as cylindrical bands; ellipsoids are enlarged for display.}
  \label{fig:overview}
\end{figure*}

\paragraph{From video to 3D and 4D scenes}
Per-scene optimization fits Gaussians to recorded driving videos~\cite{streetgaussians,omnire}, to generated videos~\cite{magicdrive3d,holotime}, or to recordings augmented with generated views~\cite{drivedreamer4d,recondreamer}.
FreeVS~\cite{freevs} instead synthesizes novel trajectories with conditional video diffusion and no explicit geometry.
Feed-forward methods predict Gaussians in one pass from pixel-aligned features~\cite{pixelsplat,mvsplat,depthsplat}, surround-view inputs~\cite{drivingforward}, pixel--voxel representations~\cite{omniscene} or volumes~\cite{evolsplat}.
Driving-specific models extend VGGT to 4D~\cite{recondrive}, enforce photometric and geometric consistency across views~\cite{mcsplat} or aggregate frames through Gaussian velocities~\cite{storm}.
Some generative world models output such scenes directly.
WorldSplat~\cite{worldsplat} decodes multi-view latents into pixel-aligned Gaussians and accumulates static geometry, DiST-4D~\cite{dist4d} generates metric depth with the video, and X-Scene~\cite{xscene} and Drive2Gauss~\cite{drive2gauss} lift generated scenes into 3D or 4D Gaussians.
Reconstructed panoramas~\cite{pansplat,matrix3d} are equirectangular images with one centre each.
In contrast, our decoder lifts panoramic latents along the rays of the six source cameras.

\section{Method}
\label{sec:method}

PanDWM takes recorded panoramic history, 3D box and HD-map layouts and a text prompt, and predicts future panoramic video that can be lifted into renderable Gaussian scenes (\cref{fig:overview}).
The panoramic band places every camera boundary inside one image (\cref{sec:method:pano}).
A block-causal diffusion model generates the band window by window (\cref{sec:method:causal}), a four-step student makes this generation fast and robust to its own history (\cref{sec:method:distill}), and a feed-forward decoder lifts each frame into 3D and 4D Gaussians (\cref{sec:method:gs}).

\subsection{Panoramic band representation}
\label{sec:method:pano}

Six separately generated views have to agree at every camera boundary.
We avoid such boundaries by generating one $360^\circ\times33.3^\circ$ panoramic band per frame.
We stitch the six nuScenes images onto one azimuth--elevation grid with the projection of OneBEV~\cite{onebev}, which gives every pixel a single source camera, and keep only the central $33.3^\circ$ of elevation.
The band is split into two overlapping halves of $2304\times256$, which a multi-view video backbone processes as two views per frame.
Both cuts lie inside single cameras, the blended overlap in the front camera and the wrap-around in the rear one.
Every camera boundary therefore lies within one half, and consistency across it becomes spatial continuity within a single view.

\subsection{Causal panoramic video diffusion}
\label{sec:method:causal}

\paragraph{Backbone and conditioning}
Each half is encoded by the Stable Diffusion 3.5 VAE.
The denoiser is the diffusion transformer of Stable Diffusion 3 (SD3)~\cite{sd3} with UniMLVG's cross-view and temporal extensions~\cite{unimlvg}, initialized from UniMLVG weights.
To let the halves exchange information along the full $360^\circ$ row at low cost, we use row-wise cross-half attention, in which each token row attends to the same row of both halves.
Temporal attention runs over the window separately for each half.
Class-coloured 3D box wireframes and HD-map outlines are stitched into the band like the images and enter through image-adapter residuals, while CLIP and T5 embeddings of a short text prompt enter SD3's joint attention.
Fourier-encoded rays of a virtual pinhole camera facing each half's centre condition the cross-view and temporal blocks~\cite{unimlvg} and identify each half and its heading.

\paragraph{Causality and teacher training}
Autoregressive rollout requires that no frame sees the future.
Each window contains 3 clean history frames and 16 target frames, and a frame-block causal mask~\cite{longlive} lets each frame attend only to itself and earlier frames.
The backbone has no temporal convolutions, and the mask alone determines causality.
The history is excluded from the loss, and all target frames share one noise level in rectified-flow training~\cite{sd3}.
The teacher samples with 50 guided Euler steps, and the last 3 output frames of each window become the history of the next.

\subsection{Four-step student trained on its own history}
\label{sec:method:distill}

Fifty guided steps cost 100 denoiser evaluations per window, and a model trained on recorded history meets its own imperfect outputs as history at inference.
We address both with a four-step student trained on self-generated history.

\paragraph{Initialization and sampling}
The causal student $f_\theta$ predicts clean latents and is initialized from the teacher by causal consistency distillation~\cite{cm,causalforcingpp}.
It minimizes $\|f_\theta(x_{\sigma_n})-\operatorname{sg}[f_{\bar\theta}(\tilde{x}_{\sigma_{n+1}})]\|^2$, where $\tilde{x}_{\sigma_{n+1}}$ results from one guided teacher Euler step from $\sigma_n$ to $\sigma_{n+1}$, $\bar\theta$ is the moving average of $\theta$ and $\operatorname{sg}$ stops gradients.
The student then generates each window in four unguided steps at $\sigma\in\{1,0.9,0.75,0.5\}$, predicting a clean latent and re-noising it between steps.

\paragraph{Training on self-generated history}
Starting from 3 recorded frames, the student rolls out windows without gradients, and the final 3 generated frames become the history of the training window~\cite{selfforcing,causaldrive}.
Gradients flow through a randomly selected denoising exit step of that window, and the history latents stay clean.
DMD~\cite{dmd,dmd2} matches the student's output distribution to the teacher's.
It uses the frozen guided teacher and a fake-score network trained on student samples, both conditioned on the student's history.

\paragraph{Adversarial supervision}
DMD pulls the student toward its teacher rather than toward recorded video.
To anchor the student to recordings, we add spatial and temporal discriminators on intermediate fake-score features.
Spatial heads score pooled tokens, and temporal heads score consecutive-frame differences.
A hinge loss compares generated windows with recordings at the same time steps and noise level, including the first window after recorded history.

\paragraph{Motion and sharpness}
We further match coarse motion, flicker and detail to the recorded latents.
Let $\mathcal{P}$ denote average pooling and $\mathcal{H}x=x-\mathcal{U}\mathcal{P}x$ the high-frequency residual, with $\mathcal{U}$ upsampling to the latent resolution.
We compute three generated-to-recorded ratios: $\rho_{\text{lf}}$ of the mean squared frame difference of $\mathcal{P}x$, $\rho_{\text{hf}}$ of the frame difference of $\mathcal{H}x$ normalized by its energy, and $\rho_{\text{e}}$ of the high-frequency energy.
The losses are
\begin{align}
  \mathcal{L}_{\text{mm}}&=[\log \rho_{\text{hf}}]_+^2+[\log \rho_{\text{lf}}]_+^2+[\log 0.9-\log \rho_{\text{lf}}]_+^2,
  \label{eq:motion}\\
  \mathcal{L}_{\text{sharp}}&=[\log 0.9-\log \rho_{\text{e}}]_+^2,
  \label{eq:sharp}
\end{align}
where $[\cdot]_+=\max(\cdot,0)$.
The low-frequency term keeps coarse motion within 0.9--1 times the recordings, the high-frequency term penalizes flicker and the energy term discourages blur.
Frame differences include the transition from the last history frame to the first generated frame, which keeps consecutive windows continuous.
Four successive training stages add DMD with the spatial heads, the temporal heads, $\mathcal{L}_{\text{mm}}$ and $\mathcal{L}_{\text{sharp}}$.

\subsection{Feed-forward Gaussian lifting}
\label{sec:method:gs}

\begin{figure}[t]
  \centering
  \includegraphics[width=\linewidth]{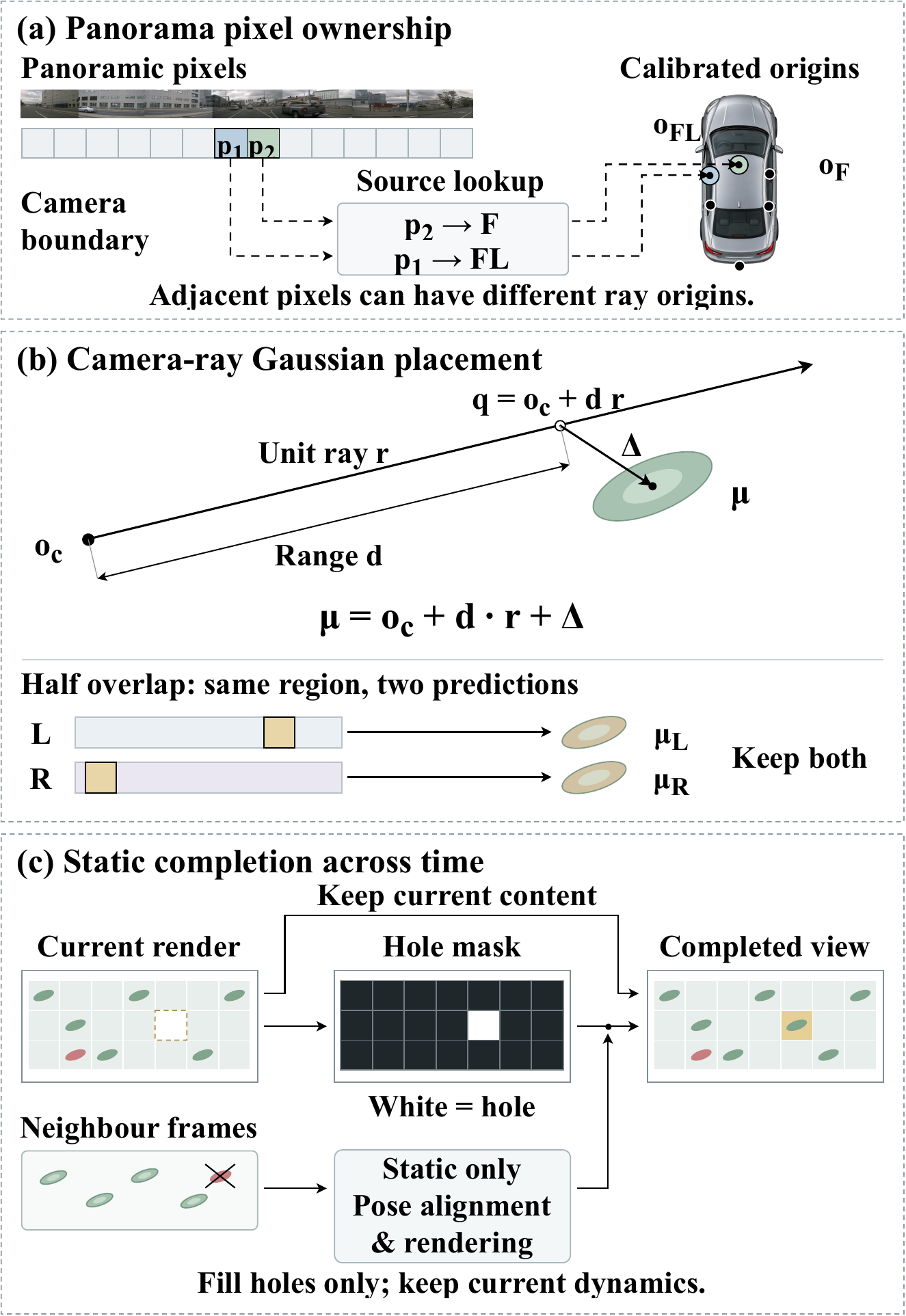}
  \caption{Gaussian geometry and completion.
  (a) Panoramic pixels retain their calibrated source-camera origins.
  (b) Range and offset locate Gaussians relative to these rays (\cref{eq:gsmean}); both halves' predictions are retained in the overlap.
  (c) Ego-motion-aligned neighbouring static Gaussians fill uncovered pixels, while dynamic content comes from the current frame.}
  \label{fig:geometry}
\end{figure}

\paragraph{Decoder and geometry}
Adjacent pixels of the band can come from cameras with different optical centres.
We therefore place each Gaussian on its own source-camera ray.
We adapt the pixel-aligned Gaussian decoder of WorldSplat~\cite{worldsplat} to the band and train it from scratch.
The decoder takes the panoramic latents and Pl\"ucker ray maps of the calibrated source-camera rays.
Cross-half attention and convolutional upsampling then predict one Gaussian per pixel, with colour, opacity, scale, rotation, range $d_p$ and offset $\Delta_p$.
With $c(p)$ the source camera of pixel $p$, the Gaussian centre is
\begin{equation}
  \mu_p=o_{c(p)}+d_p\,r_p+\Delta_p,
  \label{eq:gsmean}
\end{equation}
where $o_{c(p)}$ is that camera's optical centre, $r_p$ the calibrated unit ray of $p$, $d_p\in[0.5,400]$\,m a range predicted in inverse depth, and $\Delta_p$ a learned offset across the ray, bounded to one decoder pixel.
Unlike the virtual-camera rays that condition the video model, these rays keep the distinct source-camera origins.
Both halves' Gaussians are kept in the overlap (\cref{fig:geometry}).

\paragraph{Supervision and depth relabelling}
The Gaussians are rendered into the six nuScenes cameras, and the renders are differentiably stitched back into the band for panoramic L1, LPIPS and auxiliary depth losses~\cite{worldsplat}.
Part of the 3D decoder's training uses student-generated latents, with their VAE decodings as targets and no depth loss, so that the decoder also learns to lift predicted content.
To improve the predicted ranges, each Gaussian is moved along its ray to the range that Depth Anything 3~\cite{da3} estimates from the input frames, jointly over the clip for 4D Gaussians.
Pixels that Depth Anything 3 labels as sky or leaves without a range keep $d_p$.
The decoder is fine-tuned with this relabelling so that its other attributes fit the new geometry.

\paragraph{4D Gaussians}
A single frame leaves regions unobserved along new trajectories.
The conditioning 3D boxes separate dynamic from static Gaussians.
The 4D decoder starts from the 3D one and is trained on multiple frames, whose static Gaussians are aligned by ego motion and combined with the target frame's dynamic Gaussians.
Leave-one-out reconstruction encourages complementary geometry, and an opacity coverage term encourages complete renders.
At render time, aligned static Gaussians from neighbouring frames fill pixels that the current frame leaves uncovered, and the current frame keeps its dynamic content.
For recorded input, only earlier frames are used.

\section{Experiments}
\label{sec:exp}

\subsection{Setups}
\label{sec:exp:setup}

\paragraph{Data}
We train on the 700 training scenes of nuScenes~\cite{nuscenes} and evaluate on its 150 validation scenes.
Annotated 3D boxes and HD maps are rendered into panoramic layout images as conditions (\cref{sec:method:pano}).
Val-150 contains the first 3.5\,s of each validation scene, following the evaluation configuration of UniMLVG~\cite{unimlvg} in the OpenDWM repository~\cite{opendwm}, and serves as the benchmark for comparison with other methods.
For longer horizons we use the 83 validation scenes that allow a 19.5\,s rollout.
Test-S takes three clips from each, starting 5, 10 and 15\,s in (249 clips), and Test-L scores windows 3--12 of a 19.5\,s rollout from each scene start.
Both test sets are disjoint in time from Val-150.
Ablations use the first 3.5\,s of 78 validation scenes (Dev-78), and checkpoints were selected on Dev-78 and on long rollouts of 12 of the 83 scenes.
Val-150, Test-S and Dev-78 clips have 3 recorded history frames followed by 32 generated frames.

\paragraph{Metrics}
Video quality is measured with FID~\cite{fid} and I3D-based~\cite{i3d} FVD~\cite{fvd}, where FVD-1 and FVD-2 cover the first and second 16-frame windows.
The flicker ratio compares frame-to-frame changes of generated and recorded video on static scenes, with 1 as the reference, and the sharpness ratio compares their detail.
Controllability is measured by object recall from Faster R-CNN~\cite{fasterrcnn} and road IoU from Mask2Former~\cite{mask2former}, and efficiency by denoiser evaluations (NFE) and time per frame.
All methods are compared with equal sample counts and identical valid regions, alongside real-vs-real reference distances~\cite{chong2020fid} and temporal measures~\cite{fvdbias}.
For our method we report means and standard deviations over three seeds, with 95\% scene-level jackknife intervals.

\paragraph{Evaluation spaces and baselines}
We run UniMLVG~\cite{unimlvg}, the SD3.5-based OpenDWM model~\cite{opendwm}, DreamForge~\cite{dreamforge}, DiST-4D~\cite{dist4d} and OmniNWM~\cite{omninwm} with official weights and sampling settings (reproduction checks in supplementary~\cref{sec:suppl:baselines}).
All of them predict from recorded history and are scored on the same 32 target frames of each Val-150 clip, with 12\,Hz outputs matched to these frames by LiDAR timestamp.
DiST-4D and OmniNWM take a single recorded frame as history, and OmniNWM receives no boxes or map.
Quality is evaluated in the $4114\times256$ panoramic band that our model generates, into which baseline outputs are stitched with the OneBEV~\cite{onebev} mapping.
For a second evaluation, our band is mapped back into the six real camera views at $256\times448$.
Gaussian baselines are DrivingForward~\cite{drivingforward}, Omni-Scene~\cite{omniscene}, ReconDrive~\cite{recondrive} and MCSplat~\cite{mcsplat}.

\paragraph{Implementation details}
We train in four stages on 8 GPUs with AdamW, with learning rates of $2\times10^{-6}$ for the student, $4\times10^{-7}$ for the fake score and $3\times10^{-5}$ for the discriminator heads, and an EMA decay of 0.99; the stages and other settings are in supplementary~\cref{sec:suppl:student}.

\subsection{Comparison with driving video models}
\label{sec:exp:main}

\begin{table*}[t]
  \caption{Comparison with driving video models on Val-150, evaluated in the panoramic band. Hist.: number of recorded history frames. FVD-1 and FVD-2 cover the first and second 16-frame windows. NFE counts denoiser evaluations per window, including both guidance branches. Time is measured on one A40 including VAE decoding, per 0.1\,s of video (one frame at 10\,Hz; DreamForge, DiST-4D and OmniNWM output 12\,Hz). PanDWM is averaged over three seeds and baselines use seed 42. Bold marks the best method; $^\dagger$the causal teacher of PanDWM is a reference and is not ranked.}
  \label{tab:main}
  \centering
  \setlength{\tabcolsep}{3pt}
  \begin{tabular}{@{}lccccccccc@{}}
    \toprule
    Method & Views & Hist. & NFE & s/frame & FID$\downarrow$ & FVD-1$\downarrow$ & FVD-2$\downarrow$ & Obj.\ recall$\uparrow$ & Road IoU$\uparrow$ \\
    \midrule
    UniMLVG~\cite{unimlvg} & 6 & 3 & 100 & 20.3 & 24.3 & 51.3 & 64.5 & \textbf{66.8} & 74.0 \\
    OpenDWM (SD3.5)~\cite{opendwm} & 6 & 3 & 80 & 22.9 & 30.5 & 60.3 & 96.6 & 64.8 & \textbf{75.7} \\
    DreamForge~\cite{dreamforge} & 6 & 3 & 60 & 35.9 & 17.3 & 46.0 & 64.8 & 48.3 & 72.4 \\
    DiST-4D~\cite{dist4d} & 6 & 1 & 60 & 15.9 & 17.2 & 55.8 & 71.0 & 41.4 & 72.2 \\
    OmniNWM~\cite{omninwm} & 6 & 1 & 50 & 89.9 & 25.4 & 73.0 & 113.9 & 30.4 & 71.3 \\
    \midrule
    Causal teacher$^\dagger$ & 2 & 3 & 100 & 33.7 & 9.93 & 47.7 & 82.5 & 58.9 & 75.4 \\
    PanDWM (Ours) & 2 & 3 & \textbf{4} & \textbf{1.96} & \textbf{10.1} & \textbf{37.4} & \textbf{62.1} & 62.6 & 75.5 \\
    \bottomrule
  \end{tabular}
\end{table*}

\begin{figure}[t]
  \centering
  \includegraphics[width=\linewidth]{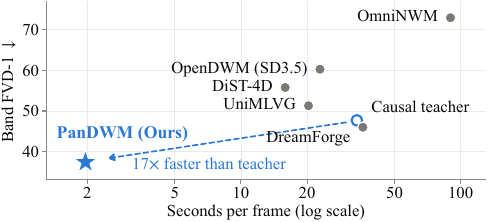}
  \caption{Band FVD-1 against time per frame on Val-150 (FID in \cref{tab:main}); lower left is better.}
  \label{fig:quality_speed}
\end{figure}

\paragraph{Short-term prediction}
PanDWM has a lower FID, FVD-1 and FVD-2 in the band than every baseline and generates 8 to 46 times faster, at 1.96\,s per frame (\cref{tab:main,fig:quality_speed}).
It also has higher object recall and road IoU than DreamForge, DiST-4D and OmniNWM.
Generating the band natively avoids the resampling loss of six-view outputs.
Recorded views at $256\times448$ already score 17.9 FID once stitched into the band, whereas mapping the band into the six cameras costs 0.63 FID on recorded data.
In the six cameras, PanDWM also has the lowest FID (5.34).
At camera boundaries it adds no seams beyond those of recorded video, unlike four of the five baselines (supplementary~\cref{sec:suppl:seams}).
Confidence intervals and the full six-camera evaluation are in supplementary~\cref{sec:suppl:val150_stats}.

\paragraph{Front-view world models}
We also compare the front view of our panoramas with HorizonDrive~\cite{horizondrive}, Epona~\cite{epona}, Vista~\cite{vista} and GEM~\cite{gem} on 149 of the Val-150 clips (\cref{fig:frontview}; settings in supplementary~\cref{sec:suppl:baselines}).
The data loader used for these runs excludes scene-0018, in which one frame is 50.5\,ms off, beyond its 50\,ms synchronization tolerance (Val-150 uses 51\,ms).
At 2.5\,s, PanDWM keeps the marked objects in place, whereas the front-view models lose or displace them.

\begin{figure*}[t]
  \centering
  \includegraphics[width=\textwidth]{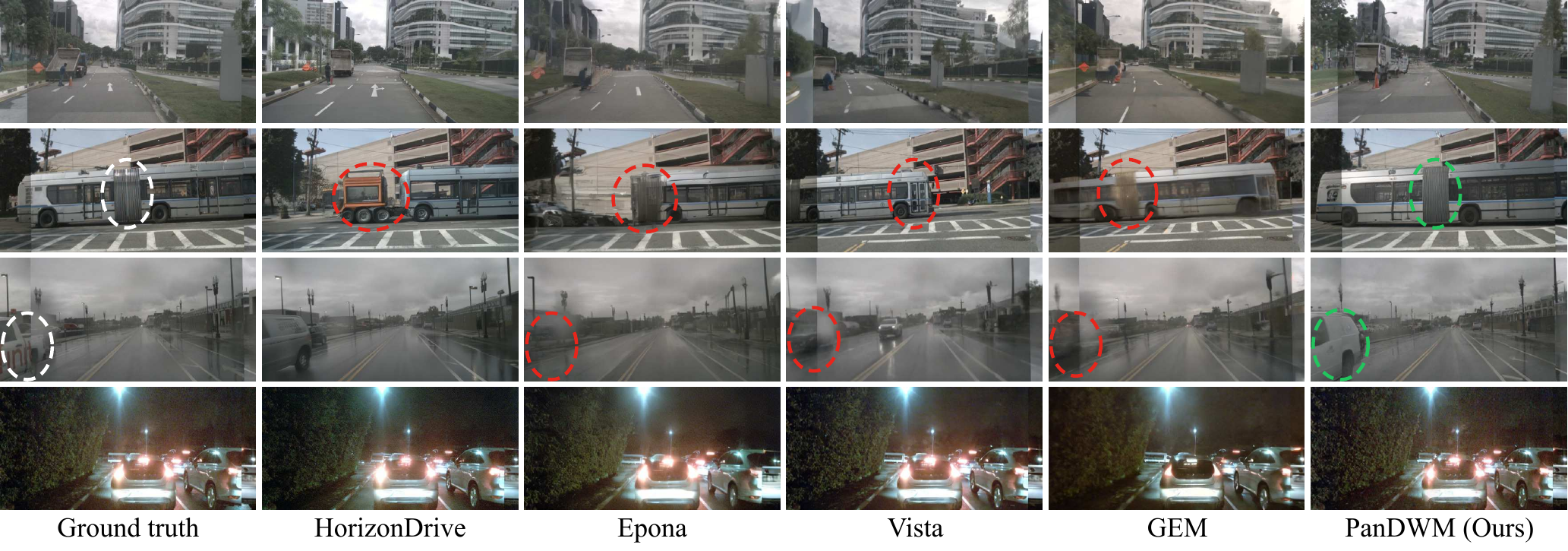}
  \caption{Front-view predictions at 2.5\,s for four fixed validation clips. Dashed ellipses mark corresponding objects (white: ground truth; green: kept; red: lost or displaced). PanDWM and HorizonDrive use layouts, Epona uses ego trajectory, and Vista and GEM use image history alone.}
  \label{fig:frontview}
\end{figure*}

\begin{figure*}[t]
  \centering
  \includegraphics[width=\textwidth]{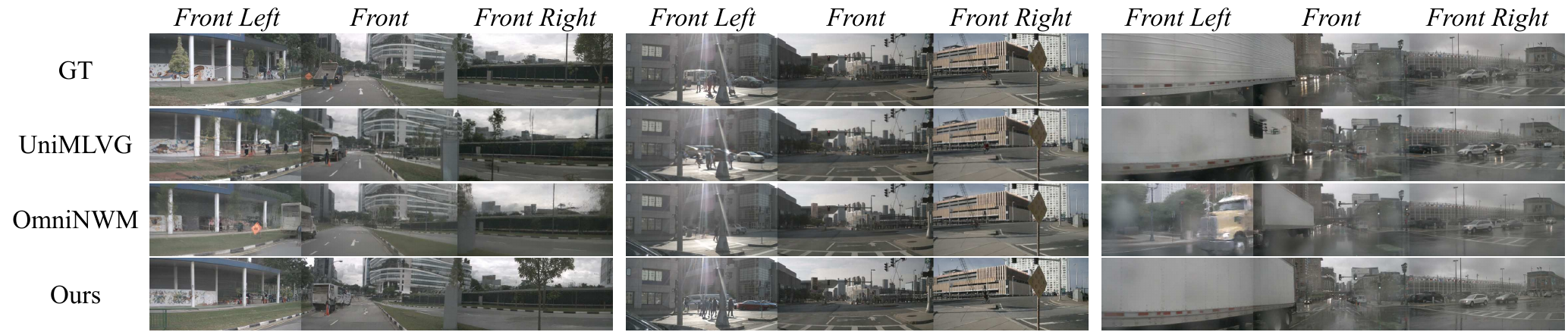}
  \caption{Front-left, front and front-right panoramic sectors at 2.5\,s for three fixed validation clips. Ground truth and PanDWM are native panoramas; UniMLVG and OmniNWM are mapped from six views. Crops cover about $\pm13^\circ$ elevation; brightness steps at camera boundaries come from the recorded exposures, which PanDWM reproduces. All methods use seed 42; OmniNWM receives one recorded frame and the ego trajectory.}
  \label{fig:multiview}
\end{figure*}

\paragraph{Multi-camera comparison}
\cref{fig:multiview} compares our panoramas with those of UniMLVG and OmniNWM~\cite{omninwm}, whose six views are stitched with the OneBEV mapping of \cref{tab:main}.
Across camera boundaries, PanDWM reproduces the appearance of the recorded band.

\subsection{Comparison with the teacher and efficiency}
\label{sec:exp:ladder}
\cref{tab:main} also lists the causal teacher from which PanDWM is distilled, which samples with 50 guided steps, \ie 100 denoiser evaluations.
With 25$\times$ fewer evaluations, PanDWM keeps the teacher's FID on Val-150 ($+0.16$, $[-0.3, 0.6]$) and lowers FVD-1 ($-10.4$, $[-13.6, -7.2]$) and FVD-2 ($-20.4$, $[-26.7, -14.0]$).
The six-camera evaluation gives the same result.
The student also recovers more of the annotated objects ($+3.7$, $[2.6, 4.8]$) at the same road IoU and flickers less on static scenes (1.49 against 1.85, a difference of $-0.36$, $[-0.47, -0.25]$).
On Test-S, FVD-1 falls from 36.6 to 27.3 ($-9.3$, $[-12.2, -6.4]$) and FVD-2 from 64.9 to 48.5 ($-16.4$, $[-23.1, -9.7]$), with an unchanged FID (7.85 against 7.81).
Even without key-value caching, PanDWM runs 17$\times$ faster than its teacher (1.96\,s against 33.7\,s per frame).
Each window update recomputes all 19 frames and commits 16, \ie 0.25 denoiser evaluations per generated frame against the teacher's 6.25.

\subsection{Ablations}
\label{sec:exp:ablation}

Each variant in \cref{tab:ablation} changes one component, is trained from the consistency-distilled student for 500 iterations with the same data order and seed, and is evaluated on Dev-78 with two seeds.

\begin{table}[t]
  \caption{Ablations on Dev-78 with two seeds. Each variant changes one component of the full recipe and, like it, is trained for 500 iterations from the starting student. Our model, trained longer in stages, is a reference. Best values are bold; paired differences with 95\% intervals are in the text.}
  \label{tab:ablation}
  \centering
  \setlength{\tabcolsep}{2.5pt}
  \begin{tabular}{@{}lrrrr@{}}
    \toprule
    Variant & FID$\downarrow$ & FVD-1$\downarrow$ & FVD-2$\downarrow$ & Flicker \\
    \midrule
    Starting student & 19.30 & 103.4 & 172.4 & 2.334 \\
    \midrule
    Full recipe (500 it.) & 17.07 & 77.1 & 131.3 & 2.005 \\
    w/o adversarial loss & 17.95 & 81.2 & 141.2 & 1.993 \\
    w/o temporal disc.\ heads & 16.75 & 78.0 & 131.8 & 2.041 \\
    Recorded history & 17.17 & 74.3 & 133.2 & 2.007 \\
    \midrule
    Our model & \textbf{14.28} & \textbf{56.8} & \textbf{93.3} & \textbf{1.580} \\
    \bottomrule
  \end{tabular}
\end{table}

\paragraph{Training recipe}
The full recipe improves on the starting student in all four metrics.
Our model, trained longer in stages, lowers FID by a further 2.8, FVD-1 by 20.3, FVD-2 by 38.0 and flicker by 0.42, all significant.

\paragraph{Adversarial loss}
Removing it raises FID by 0.88 ($[0.59, 1.17]$), FVD-1 by 4.0 ($[2.4, 5.7]$) and FVD-2 by 9.9 ($[6.0, 13.9]$).

\paragraph{Temporal heads}
They trade 0.32 per-frame FID ($[0.17, 0.48]$) for temporal stability: without them, flicker rises by 0.037 ($[0.018, 0.055]$).

\paragraph{Self-generated history}
Training on recorded history instead raises Test-L FVD by 26.1 over windows 9--11 ($[12.6, 39.6]$, supplementary~\cref{sec:suppl:selfhist}).
It fits the first window slightly better (FVD-1 $-2.8$, $[-4.7, -0.9]$), since the evaluated clips also start from recorded frames.

\subsection{Long-horizon generation}
\label{sec:exp:long}

\begin{figure}[t]
  \centering
  \includegraphics[width=\linewidth]{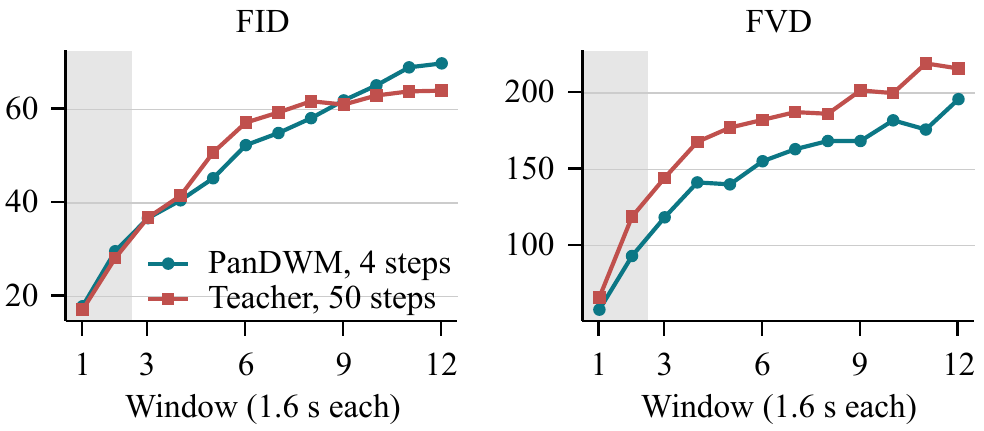}
  \caption{Per-window FID and FVD on Test-L (83 scenes, seed 42), each computed on the 83 windows of one position.}
  \label{fig:long}
\end{figure}

Over 12 autoregressive windows (19.5\,s) on Test-L, the four-step student keeps its FVD advantage over the 50-step teacher (\cref{fig:long}), with the same scenes, seed and rollout settings.

Pooled over windows 3--5, 6--8 and 9--11, the student's FVD is 86.6, 113.4 and 126.6 against the teacher's 116.7, 136.9 and 157.9 (17.2\%--25.8\% lower), and its FID remains comparable (26.1, 38.9 and 49.0 against 27.8, 43.0 and 45.9).
The teacher over-sharpens as the rollout proceeds: its sharpness ratio to the recordings rises from 0.99 to 1.54 over the 12 windows, whereas the student's moves from 0.94 to 1.04.
Against autoregressive baselines on the same rollouts, PanDWM has lower FID than UniMLVG over windows 3--8 at a tenth of its cost, and lower FID and FVD than OpenDWM in every window group (supplementary~\cref{sec:suppl:longbase}).

\begin{table}[t]
  \caption{Reconstruction from recorded images on Test-S, evaluated in the six cameras within the band. PSNR, SSIM~\cite{ssim} and LPIPS~\cite{lpips} use renders composited over gray; AbsRel and $\delta_1$ use LiDAR depth at static points within 80\,m with accumulated opacity $\geq0.5$. Best per resolution in bold; $^\ddagger$LiDAR input. Ours (4D) adds four recorded frames 0.5--2\,s before the input to the single-frame Ours (3D).}
  \label{tab:gs_recon}
  \centering
  \small
  \setlength{\tabcolsep}{2pt}
  \begin{tabular}{@{}lccccc@{}}
    \toprule
    Method & PSNR$\uparrow$ & SSIM$\uparrow$ & LPIPS$\downarrow$ & AbsRel$\downarrow$ & $\delta_1\uparrow$ \\
    \midrule
    \multicolumn{6}{@{}l@{}}{Resolution: $224\times400$} \\
    Omni-Scene~\cite{omniscene} & \textbf{21.00} & \textbf{0.597} & \textbf{0.363} & 0.333 & \textbf{0.734} \\
    MCSplat$^\ddagger$~\cite{mcsplat} & 18.85 & 0.511 & 0.442 & \textbf{0.270} & 0.644 \\
    Ours (3D) & 18.21 & 0.510 & 0.436 & 0.334 & 0.560 \\
    Ours (4D) & 19.29 & 0.533 & 0.410 & 0.298 & 0.642 \\
    \midrule
    \multicolumn{6}{@{}l@{}}{Resolution: $280\times518$} \\
    ReconDrive~\cite{recondrive} & 18.06 & 0.510 & \textbf{0.438} & 0.663 & \textbf{0.680} \\
    Ours (3D) & 18.14 & 0.516 & 0.462 & 0.334 & 0.560 \\
    Ours (4D) & \textbf{19.18} & \textbf{0.534} & \textbf{0.438} & \textbf{0.298} & 0.642 \\
    \midrule
    \multicolumn{6}{@{}l@{}}{Resolution: $352\times640$} \\
    DrivingForward~\cite{drivingforward} & 18.54 & 0.533 & \textbf{0.431} & 0.368 & 0.595 \\
    Ours (3D) & 18.09 & 0.526 & 0.486 & 0.334 & 0.561 \\
    Ours (4D) & \textbf{19.09} & \textbf{0.539} & 0.464 & \textbf{0.299} & \textbf{0.642} \\
    \bottomrule
  \end{tabular}
\end{table}

\subsection{3D and 4D Gaussian evaluation}
\label{sec:exp:gs}

We render every Gaussian scene into the six calibrated nuScenes cameras and evaluate the rows covered by the band, with the sky masked.
In \cref{tab:gs_recon}, each method reconstructs a recorded Test-S clip and renders the next frame, the frame 0.5\,s later and the frames $\pm3.2$\,m along the trajectory, twice the distance of the Omni-Scene protocol~\cite{omniscene}.
Each baseline runs at its own resolution, and ours runs at each of them.
Our Gaussians use the relabelling of \cref{sec:method:gs}, and an extra layer of Gaussians beneath the scene fills only pixels that the scene leaves uncovered (supplementary~\cref{sec:suppl:gseval}).

Our 4D Gaussians have the best PSNR and SSIM at $280\times518$ and $352\times640$ and the lowest AbsRel among camera-only methods at every resolution.
At $280\times518$ they exceed ReconDrive by 1.1\,dB in PSNR with less than half its AbsRel (0.298 against 0.663).
At $352\times640$ they exceed DrivingForward by 0.55\,dB in PSNR and also in SSIM, AbsRel and $\delta_1$.
4D completion adds 1.0--1.1\,dB over our single-frame 3D Gaussians, and lifting with relabelling takes about 1\,s per frame on one H100 (supplementary~\cref{sec:suppl:gseval}).

\paragraph{Generated scenes}
Generated videos have no ground truth at novel viewpoints.
Following WorldSplat~\cite{worldsplat} and DiST-4D~\cite{dist4d}, we score laterally shifted views by FID against recorded frames cropped to the same rows, and static-region depth by AbsRel against LiDAR (\cref{tab:gs_gen}).
We use the 83 Test-S clips starting 10\,s into each scene.
Here depth comes from the generated frames for 88\% of the pixels, and no Gaussians are added beneath the scene.
Our renders have lower FID than DiST-4D up to $\pm2$\,m and than ReconDrive applied to DiST-4D's videos at every shift.
DiST-4D synthesizes each shifted view with a novel-view diffusion model, whereas every shift of ours is rendered from one explicit 4D scene.
Refining our renders with a diffusion model as well brings FID below DiST-4D's at every shift (34.9 against 48.4 at $\pm4$\,m).
The refiner is a copy of our causal teacher, conditioned on the rendered latents through a zero-initialized branch and trained to restore recorded frames from renders.
It trades some cross-shift consistency for per-view fidelity (supplementary~\cref{sec:suppl:gseval}).
Our scenes also have a lower AbsRel than DiST-4D (0.316 against 0.467), including after median scaling (0.304 against 0.314).

\begin{table}[t]
  \caption{FID$\downarrow$ of views shifted laterally by the distance in each column, and static-region AbsRel, on 83 Test-S clips. Refinement changes the rendered images but not the Gaussians, so our two rows share one AbsRel.}
  \label{tab:gs_gen}
  \centering
  \setlength{\tabcolsep}{2pt}
  \begin{tabular}{@{}lccccc@{}}
    \toprule
    Method & 0\,m & $\pm1$\,m & $\pm2$\,m & $\pm4$\,m & AbsRel$\downarrow$ \\
    \midrule
    DiST-4D~\cite{dist4d} & 23.5 & 33.1 & 38.8 & 48.4 & 0.467 \\
    DiST-4D + ReconDrive & 39.2 & 44.5 & 57.4 & 83.7 & 0.715 \\
    Ours (4D) & 21.8 & 27.0 & 34.0 & 50.5 & \textbf{0.316} \\
    Ours (4D) + refinement & \textbf{18.5} & \textbf{19.1} & \textbf{22.8} & \textbf{34.9} & \textbf{0.316} \\
    \bottomrule
  \end{tabular}
\end{table}
\section{Conclusion}
\label{sec:conclusion}

We presented PanDWM, a driving world model that generates the surround view as a single 360$^\circ$ panoramic band.
A four-step block-causal student, trained on self-generated history, generates the band window by window, and a feed-forward decoder lifts the generated latents along calibrated source-camera rays into 3D Gaussians, which static Gaussians from neighbouring frames complete into 4D scenes.

\paragraph{Limitations and Future Work}
(1) Each window sees only the last three generated frames, and Test-L FVD grows from 86.6 (windows 3--5) to 126.6 (windows 9--11). Renders of the accumulated static Gaussians could serve as longer memory.
(2) 4D completion covers only static content, leaving unseen parts of moving objects empty; aligning them with the 3D boxes is left for future work.

\clearpage
\section*{LLM Usage Statement}
We used Claude (Anthropic), through Claude Code, and Codex (OpenAI) to assist with research implementation and manuscript preparation.
Their use included developing and debugging parts of the training, evaluation and baseline code; submitting and monitoring experiments and summarizing logged outputs; literature search and reference checking; and drafting, translating and revising text throughout the manuscript, as well as preparing figures and tables in the paper and supplementary.
The authors directed the work and take responsibility for verifying the tool-assisted content and for all scientific claims, experimental results and references.

{   
    \small
    \bibliographystyle{ieeenat_fullname}
    \bibliography{main}
}

\par

\clearpage
\setcounter{page}{1}
\maketitlesupplementary

\setcounter{section}{0}
\setcounter{figure}{0}
\setcounter{table}{0}
\setcounter{equation}{0}
\renewcommand{\theHsection}{supp.\arabic{section}}
\renewcommand{\theHfigure}{supp.\arabic{figure}}
\renewcommand{\theHtable}{supp.\arabic{table}}
\renewcommand{\theHequation}{supp.\arabic{equation}}
\renewcommand{\thefigure}{S\arabic{figure}}
\renewcommand{\thetable}{S\arabic{table}}
\renewcommand{\theequation}{S\arabic{equation}}

\makeatletter
\setlength{\@fptop}{0pt}
\setlength{\@fpsep}{12pt}
\setlength{\@fpbot}{0pt plus 1fil}
\setlength{\@dblfptop}{0pt}
\setlength{\@dblfpsep}{12pt}
\setlength{\@dblfpbot}{0pt plus 1fil}
\makeatother
\renewcommand{\floatpagefraction}{0.5}
\renewcommand{\dblfloatpagefraction}{0.5}

\section{Implementation details}
\label{sec:suppl:impl}
\subsection{Panoramic preprocessing and conditions}
\label{sec:suppl:pano}
Drawing on OneBEV~\cite{onebev}, the six nuScenes images are mapped to a $9600\times900$ canvas spanning $360^\circ$ in azimuth and $\pm25^\circ$ in elevation.
Keeping rows $[150,750)$ gives a $9600\times600$ band with $33.3^\circ$ vertical coverage.
The halves use columns $[0,5376)$ and $[4224,9600)$, each spanning $201.6^\circ$.
Their 1152-column overlap ($43.2^\circ$) lies inside the front-camera region.
Each half is resized to $2304\times256$.
For display and evaluation, removing the right half's 494 overlapping columns gives a $4114\times256$ band, with a 128-column raised-cosine cross-fade replacing the hard cut.
The rear seam wraps from column 9600 to column 0 within the rear-camera region, without overlap or blending.

The two layout inputs are class-coloured 3D box wireframes and HD map outlines.
Map elements comprise drivable areas, lanes, pedestrian crossings, walkways, stop lines and car parks within 65\,m.
The condition panoramas come from the same release as the RGB panoramas; our renderer for these conditions draws each camera with its calibration and stitches the results with the RGB mapping, so that corresponding condition and RGB pixels share a source camera.

\subsection{Video architecture and causal teacher}
\label{sec:suppl:teacher}
The SD3.5 VAE produces a $32\times288$ latent with 16 channels for each half.
A patch size of 2 gives $16\times144=2304$ tokens.
The denoiser uses the 24-layer SD3 transformer~\cite{sd3}, extended and initialized from UniMLVG~\cite{unimlvg}.
Cross-view and temporal self-attention blocks follow layers 1, 5, 9, 13, 17 and 21, with learned scalar gates blending their outputs into the main stream.
Row-wise cross-view attention exchanges same-row tokens between both halves, while temporal attention operates over all window tokens of each half.
Concatenated box and map images enter an image adapter whose residuals are added to the first six transformer blocks.
CLIP and T5 encode text for SD3's joint attention.
Each half's virtual pinhole camera faces the half's centre; its Fourier-encoded ray map is added to the cross-view and temporal block inputs.
These conditioning rays differ from the calibrated source-camera rays used for Gaussian placement.

Windows contain 19 frames at 10\,Hz, with 3 clean history frames excluded from the loss and 16 targets sharing a noise level $\sigma$.
Rectified-flow training~\cite{sd3} uses
\begin{equation}
  x_\sigma=(1-\sigma)x_0+\sigma\epsilon,\qquad
  v^\star=\epsilon-x_0,
\end{equation}
where $\epsilon\sim\mathcal{N}(0,I)$ and the timestep follows a logit-normal distribution (mean 0, standard deviation 1), mapped to $\sigma$ by the shift-3 schedule.
History is omitted in 20\% of samples, half of which train single-frame generation with the temporal blocks disabled; in a further 16\%, each history frame of each half is kept with probability 0.5. Text, boxes, map and camera rays are each dropped with probability 0.2 for classifier-free guidance.
Teacher sampling uses 50 Euler steps with guidance scale 3.

\paragraph{Causality checks}
The implementation includes future-perturbation and gradient checks: changing future inputs must leave earlier outputs bit-identical, and gradients from earlier outputs to future inputs must be exactly zero.
These checks cover the frame-block attention mask and the absence of temporal convolutions.
These checks are part of the released tests; all 55 test cases passed.

\subsection{Consistency distillation and four-step sampling}
\label{sec:suppl:cd}
Following Causal Forcing++~\cite{causalforcingpp} and consistency distillation~\cite{cm}, the teacher takes an Euler step from $\sigma_n$ to $\sigma_{n+1}$ on a 48-step grid; we add guidance to this step.
Let $h$ denote the clean history, $c$ the conditions, $v_T^{\mathrm{cfg}}$ the guided teacher velocity and $f_\theta(x,\sigma;h,c)=x-\sigma v_\theta(x,\sigma;h,c)$ the student's clean prediction.
We use
\begin{align}
  \widetilde{x}_{n+1}
    &=x_{\sigma_n}+(\sigma_{n+1}-\sigma_n)
      v_T^{\mathrm{cfg}}(x_{\sigma_n},\sigma_n;h,c),\\
  \mathcal{L}_{\mathrm{CD}}
    &=\mathbb{E}\big\|f_\theta(x_{\sigma_n},\sigma_n;h,c)
      \nonumber\\[-2pt]
    &\qquad-\operatorname{sg}\big[
      f_{\bar\theta}(\widetilde{x}_{n+1},\sigma_{n+1};h,c)
      \big]\big\|^2,
\end{align}
where $\bar\theta$ is the student's exponential moving average and $\operatorname{sg}$ stops gradients.
Both networks are initialized from the causal teacher.

For four-step generation, each step predicts $\hat{x}_0=x_{\sigma_i}-\sigma_i v_\theta(x_{\sigma_i})$ and, except at the final step, re-noises it:
\begin{equation}
  \begin{aligned}
  x_{\sigma_{i+1}}&=(1-\sigma_{i+1})\hat{x}_0+\sigma_{i+1}\epsilon_i,\\
  \epsilon_i&\sim\mathcal{N}(0,I).
  \end{aligned}
  \label{eq:sampler}
\end{equation}
The final step returns $\hat{x}_0$.
Training and sampling both use $\sigma\in\{1,0.90,0.75,0.50\}$, corresponding to indices $\{1000,750,500,250\}$ of the 1000-step grid after scheduler shift 3.
Student sampling uses no classifier-free guidance.

\subsection{Training on self-generated history}
\label{sec:suppl:student}
Each training clip starts from 3 recorded frames and rolls out $k$ windows with the four-step sampler, without gradients, so that, as in Self Forcing~\cite{selfforcing}, the student is trained on its own generated history.
$k=2$ in the first 4000 iterations, and $k$ is drawn uniformly from $\{2,\ldots,10\}$ in the last 250 (\cref{tab:suppl:stages}).
The last 3 frames of the $k$-th window form the clean history of the training window.
A randomly selected denoising exit step receives gradients; history latents remain unnoised.

\paragraph{Distribution matching}
We noise the student output $\hat{x}_0$ to a random level.
A frozen teacher with guidance scale 3 predicts $\hat{x}_{\mathrm{real}}$, and a fake-score network, initialized from the teacher and trained by flow matching on student samples, predicts $\hat{x}_{\mathrm{fake}}$.
The DMD objective~\cite{dmd,dmd2} is
\begin{align}
  g&=\frac{\hat{x}_{\mathrm{fake}}-\hat{x}_{\mathrm{real}}}
    {\operatorname{mean}|\hat{x}_0-\hat{x}_{\mathrm{real}}|},\nonumber\\
  \mathcal{L}_{\mathrm{DMD}}
    &=\tfrac12\big\|\hat{x}_0-
      \operatorname{sg}(\hat{x}_0-g)\big\|^2.
  \label{eq:dmd}
\end{align}
The mean runs over the 16 target frames.
There are five fake-score updates per student update.
Both score networks receive the same clean history as the student.

\paragraph{Adversarial heads}
Lightweight heads on fake-score blocks 4, 12 and 23 provide spatial scores from pooled tokens and temporal scores from consecutive-frame feature differences.
The hinge loss uses two comparisons with recorded frames: the training window as a stand-alone clip at the same time steps, and the first generated window after recorded history.
Real and generated inputs are noised to the same level, matched to the denoising exit step.

\paragraph{Motion statistics}
Average pooling uses a $4\times4$ kernel, and the high-frequency residual subtracts the upsampled pooled latent.
The motion ratios include the last history frame followed by the 16 generated frames; the energy ratio uses the 16 target frames.
The low-frequency ratio is based on mean squared frame differences; the high-frequency ratio normalizes frame differences by residual energy.
The main-paper losses penalize low-frequency ratios outside $[0.9,1]$, excess high-frequency change, and high-frequency energy below 0.9 times the recordings.
These terms follow measurements made during training.
Among 149 of the Val-150 clips, on the third with the least motion, the student's frame-to-frame change at iteration 1000 of the first stage is 2.18 times that of the recordings, against 1.10 for VAE reconstructions of the recordings and 2.09 for the starting student; after the temporal-head stage (1600 iterations), it is 1.96.
During the stage with the motion terms, the sharpness ratio to the recordings over windows 9--12 of 19.5\,s rollouts of eight scenes fell from 1.03 at 2000 iterations to 0.87 at 4000, which led to the sharpness term of the last stage.

\paragraph{Training stages}
The student is trained for 4250 iterations (fake-score updates, i.e.\ 850 student updates) in four stages (\cref{tab:suppl:stages}).
Each stage resumes from the last checkpoint of the previous one, and the first starts from the consistency-distilled student.
The temporal heads are trained from iteration 1150 and enter the student objective at iteration 1300.

\begin{table}[t]
  \caption{Training stages of the student. Each stage adds the listed terms to the objective of the previous stage. Clip frames include the 3 recorded history frames; $k$ is the number of self-generated windows before the training window.}
  \label{tab:suppl:stages}
  \centering
  \small
  \begin{tabular}{@{}llcc@{}}
    \toprule
    Iterations & Added & Clip frames & $k$ \\
    \midrule
    0--1150 & DMD, spatial heads & 51 & 2 \\
    1150--1600 & Temporal heads & 51 & 2 \\
    1600--4000 & $\mathcal{L}_{\text{mm}}$ & 51 & 2 \\
    4000--4250 & $\mathcal{L}_{\text{sharp}}$ & 179 & 2--10 \\
    \bottomrule
  \end{tabular}
\end{table}

\paragraph{Optimization settings}
The causal teacher is trained for 20{,}000 iterations and the consistency-distilled student for 10{,}000 before the stages of \cref{tab:suppl:stages}.
In all student stages, AdamW uses $\beta=(0, 0.999)$, with weight decay 0.01 for the student and none for the fake score, and gradients are clipped at norm 10.
Each iteration uses 8 clips, one per GPU.
The adversarial terms have weight 0.01 for the student and the discriminator heads and are ramped in between iterations 100 and 200; the temporal heads enter at iteration 1300 with a 150-iteration ramp.
The EMA of the student starts at iteration 200.

\subsection{Gaussian decoder and supervision}
\label{sec:suppl:gs}
Drawing on WorldSplat's pixel-aligned head~\cite{worldsplat}, the decoder is trained from scratch for paired panoramic latents and Pl\"ucker ray maps.
It uses a convolution to 512 channels, cross-view attention over both halves before and after a convolutional middle block, and four decoder blocks that upsample by $8\times$ to each half's $2304\times256$ grid.
Each pixel predicts colour, opacity, scale, rotation, metric range and positional offset.
Range $d\in[0.5,400]$\,m uses inverse-depth parameterization.
The source-camera origin and calibrated unit ray determine its placement, with a learned offset; both halves' Gaussian sets are concatenated without removing overlap predictions.

Rendering-based supervision follows WorldSplat~\cite{worldsplat}.
We render into the six nuScenes cameras using gsplat's anti-aliased mode, then differentiably stitch the renders with the data mapping.
L1 and LPIPS supervise colour, with a weak depth term clipped at 100\,m.
The 3D stages use precomputed dense per-camera depth maps as targets; the later 4D stages, including the final fine-tuning, use Depth Anything 3~\cite{da3} estimates computed during training on the recorded panoramas.
In the last 1000 steps of the 3D stages, outputs of the consistency-distilled four-step student replace the recorded latents with probability 0.5 for 500 steps and 0.75 for 500 steps; the 4D stages use recorded latents only.
For student latents, the VAE decoding of the same latent supplies the target and depth supervision is disabled.

\subsection{4D Gaussian training and completion}
\label{sec:suppl:gs4d}
In training, Gaussians whose centres fall inside a 3D box of the eight movable classes (car, truck, bus, trailer, construction vehicle, pedestrian, motorcycle and bicycle) enlarged by 0.3\,m are dynamic; others are static.
At render time, completion and depth fusion use every annotated box, so objects such as barriers and traffic cones also keep only the Gaussians of frame $t$; the static depth metrics exclude only the eight movable classes.
The 4D decoder starts from the 3D decoder and trains on three recorded frames per clip.
Besides individual-frame reconstruction, their static Gaussians are transformed into the target frame using ego motion and rendered with the target frame's dynamic Gaussians.
A leave-one-out term renders the target from the other frames' static Gaussians.
Coverage supervision penalizes $1-\alpha$ on valid pixels, preventing transparent Gaussians from matching dark regions against a black background.
At render time, the static Gaussians of frames $t-20$, $t-15$, $t-10$, $t-5$, $t+5$ and $t+10$ that have opacity at least 0.3 and lie within 40\,m horizontally of their own frame's ego position fill pixels not covered by frame $t$.
This offline completion uses both preceding and following frames; dynamic content comes only from frame $t$.

\subsection{Gaussian evaluation settings and render refinement}
\label{sec:suppl:gseval}
\paragraph{Depth relabelling and fine-tuning}
In \cref{tab:gs_recon,tab:gs_gen}, each Gaussian is moved along its pixel ray to the range that Depth Anything 3~\cite{da3} estimates from the input frames, the recorded panoramas for reconstruction and the VAE-decoded generated frames for generated scenes; pixels labelled sky or without an estimate below 99\,m keep the decoder's range.
For 3D Gaussians the estimate uses the frame alone.
For 4D Gaussians it is computed jointly over the clip, and the depth of each static pixel is fused with the static points of frames $t\pm2$, $\pm4$, $\pm6$ and $\pm8$ moved into the same camera by ego motion: candidates within 10\% of the pixel's own depth are accepted and the median is taken.
For reconstruction, only recorded frames up to the input frame are loaded, so the estimate and the fusion use earlier frames only, and the 4D scene adds the static Gaussians of the frames 0.5, 1, 1.5 and 2\,s before the input; no target frame and no later frame is used, and a history frame that coincides with a target is left out for that target.
The learned offset keeps only its component across the ray, softly bounded to one decoder pixel.
The decoder is fine-tuned for 2000 steps with the same relabelling inside training and an LPIPS weight of 0.5 on the frame, composite and leave-one-out renders.
With the decoder used before this fine-tuning, on 78 development clips, the per-pixel relabelling gives 4D reconstruction PSNR 18.70, static AbsRel 0.269 and $\delta_1$ 0.678 at $224\times400$, and shifted-view FID 19.9 and 46.3 at 0 and $\pm4$\,m; keeping the decoder's own ranges and correcting only one DA3 scale per frame gives 18.91, 0.330, 0.557, 18.8 and 59.6.
The relabelling therefore improves depth and large shifts at the cost of consistency near the recorded trajectory (FVD-1 at 0\,m 107.5 against 69.5).

\paragraph{Gap-filling Gaussians for reconstruction}
For \cref{tab:gs_recon}, Gaussians are added beneath the scene: they are rendered separately and composited under the current frame and the 4D fill, so they change only pixels that the scene leaves uncovered.
(i)~Seam bridges: the six cameras have different optical centres, so the scene between the last ray of one camera and the first ray of the next lies in no pixel and appears as an empty vertical stripe from other viewpoints; in every row, 16 Gaussians interpolate position, colour and opacity between the nearest kept Gaussians on the two sides of each seam (searched up to 6 pixels away), unless the gap is longer than half their range.
(ii)~Background extension: across horizontal depth edges where the range grows by more than 20\%, the first kept background Gaussian (within 6 pixels) is copied onto the rays of up to 8 foreground pixels at its own range.
(iii)~Band extension: the top and lowest kept rows continue 8 rows beyond the band, on the ground plane for ground pixels and as vertical surfaces otherwise.
On Test-S at $224\times400$, these raise coverage from 0.81 to 0.90 and PSNR from 17.03 to 18.21 for 3D Gaussians, and coverage from 0.95 to 0.97 and PSNR from 18.89 to 19.29 for 4D Gaussians, with AbsRel changing by at most 0.003.
They are not used for generated scenes, where our renders drop frame $t$'s Gaussians at depth discontinuities and those larger than 1\,m, and pixels left uncovered are filled by push-pull interpolation for every rendered method.

\paragraph{Render refinement}
The refiner is a copy of the causal teacher with one additional, zero-initialized input branch: the 16-channel latent of the rendered view (VAE posterior mean) and its coverage are patchified by a $2\times2$ convolution to the transformer width and added to the patch embedding, so that at initialization the refiner equals the teacher.
Training pairs come from 1622 recorded training clips: a Gaussian scene built from the recorded video with an earlier checkpoint of our decoder is rendered at the recorded trajectory and the recorded frames are the targets.
Each clip is degraded in one of three ways: visibility culling that mimics lateral shifts of 0.5 to 4\,m (70\%), background taken only from neighbouring frames (15\%), or none (15\%).
It is trained for 4000 steps.
At test time, the latent of the rendered shifted view, noised to $\sigma=0.8$, is denoised in 20 steps with classifier-free guidance of scale 2 on all conditions; the 35 frames are processed as two 19-frame windows, and the output, decoded and stitched as the video model's output, replaces the render.
In FVD-1 on the 83 Test-S clips, the renders reach 86.5, 122.3, 174.9 and 304.4 at 0, $\pm1$, $\pm2$ and $\pm4$\,m, the refined views 62.0, 73.8, 97.5 and 159.2, and DiST-4D 79.7, 118.5, 148.6 and 205.5.

\paragraph{Development clips}
The decoder, the relabelling and completion settings and the refinement sampler were chosen on GS-Dev, the clips starting 10\,s into the 78 Dev-78 scenes; 12 of these clips are also Test-S clips in \cref{tab:gs_recon,tab:gs_gen}.
On the clips of the other 71 scenes (213 for reconstruction, 71 for generated scenes), the best value of every reconstruction column and every comparison between our rows and DiST-4D on generated scenes stay the same.

\paragraph{Runtime}
On one H100, the decoder lifts a frame into Gaussians in 257\,ms, building the 4D scene from six already decoded neighbours takes 47\,ms, and rendering a $224\times400$ view takes 5\,ms.
The clip-level Depth Anything 3 relabelling with fusion takes 21.8\,s per 35-frame clip, i.e.\ 0.68\,s per generated frame, so the Gaussian stage costs about 1.0\,s per frame.

\paragraph{Shift preservation and multi-view consistency}
To check that refinement keeps the lateral shift, we compare $16\times16$ average-pooled views on eight Test-S clips (frames 10, 20 and 30).
At $\pm1$, $\pm2$ and $\pm4$\,m, the refined views are 28.5--28.8, 26.2--26.8 and 23.5--23.9\,dB in PSNR from the render at the same shift, but 21.1, 18.7 and 17.0\,dB from the unshifted render, close to the renders' own distance from it (21.5, 18.7--18.8 and 16.6--16.7\,dB); no frame is closer to the unshifted render.
Refinement reduces consistency across shifts.
Warping one view into another with the Gaussian geometry on the 71 Test-S clips, the $4\times4$-pooled PSNR between the unshifted view and the views at $\pm1$, $\pm2$ and $\pm4$\,m is 43.5--43.7, 39.7--40.1 and 35.9--36.2\,dB for renders but 32.7--32.9, 29.9--30.5 and 26.7--26.8\,dB for refined views.
Most of this drop is sampling randomness: refining the same view twice with different seeds gives 36.3, 34.1, 32.7 and 31.4\,dB at 0, 1, 2 and 4\,m on 24 clips, and on the same clips the cross-shift values (32.9, 30.8 and 27.7\,dB) are only 1.2, 1.9 and 3.7\,dB below this floor.

\paragraph{Figure conventions}
Main-paper image inserts come from one development scene.
Layout insets show the front region with thickened lines.
The oblique Gaussian-scene rendering uses actual 3D Gaussians; enlarged ellipsoids, cameras and rays are schematic.
Latent cards and attention masks are schematic, with each temporal-mask cell representing attention between two whole-frame token blocks.

\section{Additional results}
\label{sec:suppl:results}
\subsection{Val-150 prediction statistics}
\label{sec:suppl:val150_stats}
On Val-150, PanDWM reaches FID $10.09\pm0.08$ (95\% interval $[9.5, 10.7]$), FVD-1 $37.4\pm1.0$ ($[34.8, 40.0]$) and FVD-2 $62.1\pm2.2$ ($[57.9, 66.4]$) over three seeds.
At half the sample size, two scene-disjoint halves of the recorded clips score FID 40.2, FVD-1 143.2 and FVD-2 144.2.

\paragraph{Six camera views}
For the evaluation in the six real camera views at $256\times448$, our panoramas are resampled into the cameras with the inverse of the OneBEV mapping, and the same valid regions are used for all methods.
The conversion itself costs little: the recorded band resampled into the cameras scores FID 0.63 and FVD-1 3.6 against the recorded camera views.
\Cref{tab:suppl:cam} lists the six-camera results of the methods in \cref{tab:main}.
Against the causal teacher, PanDWM is within noise in FID ($-0.15$, $[-0.3, 0.0]$) and lower in FVD-1 ($-8.1$, $[-10.9, -5.3]$) and FVD-2 ($-13.7$, $[-19.4, -8.1]$).
Against DreamForge, which receives the same three recorded frames, PanDWM is lower in the band in FID ($-7.2$, $[-8.7, -5.8]$) and FVD-1 ($-8.6$, $[-12.9, -4.4]$), and the FVD-2 difference is not significant ($-2.7$, $[-9.4, 4.0]$); in the six cameras, PanDWM is lower in FID ($-4.4$, $[-5.0, -3.9]$), the FVD-1 difference is not significant ($-0.9$, $[-3.6, 1.8]$), and DreamForge has lower FVD-2 ($+8.8$, $[3.6, 13.9]$).
The comparisons with UniMLVG, OpenDWM (SD3.5), DiST-4D and OmniNWM follow.

\begin{table}[t]
  \caption{Six-camera evaluation on Val-150 at $256\times448$ with recorded history. PanDWM is averaged over three seeds; the other methods use seed 42. Best method values are bold; $^\dagger$the causal teacher is a reference and is not considered.}
  \label{tab:suppl:cam}
  \centering
  \small
  \begin{tabular}{@{}lccc@{}}
    \toprule
    Method & FID$\downarrow$ & FVD-1$\downarrow$ & FVD-2$\downarrow$ \\
    \midrule
    UniMLVG~\cite{unimlvg} & 5.38 & \textbf{23.4} & \textbf{38.6} \\
    OpenDWM (SD3.5)~\cite{opendwm} & 6.46 & 23.8 & 47.3 \\
    DreamForge~\cite{dreamforge} & 9.77 & 27.8 & 41.5 \\
    DiST-4D~\cite{dist4d} & 11.92 & 35.8 & 47.2 \\
    OmniNWM~\cite{omninwm} & 14.38 & 43.1 & 79.0 \\
    \midrule
    Causal teacher$^\dagger$ & 5.49 & 35.0 & 64.0 \\
    PanDWM (ours) & \textbf{5.34} & 26.9 & 50.2 \\
    \bottomrule
  \end{tabular}
\end{table}

\paragraph{Comparison with UniMLVG}
In the band, UniMLVG reaches FID 24.3 ($[22.6, 26.0]$), FVD-1 51.3 ($[47.3, 55.3]$) and FVD-2 64.5 ($[59.8, 69.1]$).
PanDWM is lower in FID by 14.2 (difference $-14.2$, $[-15.7, -12.7]$) and in FVD-1 by 13.9 ($-13.9$, $[-18.2, -9.6]$); the FVD-2 difference is not significant ($-2.4$, $[-7.7, 3.0]$).
Most of the FID difference reflects resolution: recorded views at $256\times448$, stitched in the same way, already score 17.9, and UniMLVG is 6.4 ($[5.2, 7.6]$) above them.
In the six cameras at that resolution, the FIDs do not differ significantly (PanDWM 5.34, UniMLVG 5.38; difference $-0.04$, $[-0.26, 0.18]$).
PanDWM has higher FVD-1 (26.9 against 23.4; $+3.5$, $[1.1, 5.9]$) and FVD-2 (50.2 against 38.6; $+11.7$, $[7.3, 16.1]$).
On static scenes, PanDWM changes 1.49 times as much from frame to frame as the recorded band ($[1.35, 1.62]$) and UniMLVG 1.41 ($[1.24, 1.57]$).
Relative to recorded views at its own resolution, whose ratio is 0.78, UniMLVG's ratio is 1.86 ($[1.60, 2.12]$).

\paragraph{Comparison with OpenDWM (SD3.5)}
In the band, OpenDWM (SD3.5) reaches FID 30.5 ($[28.3, 32.7]$), FVD-1 60.3 ($[55.2, 65.3]$) and FVD-2 96.6 ($[87.1, 106.1]$); PanDWM is lower in all three, with differences of $-20.4$ ($[-22.5, -18.3]$), $-22.9$ ($[-27.8, -17.9]$) and $-34.5$ ($[-43.8, -25.2]$).
In the six cameras, PanDWM is lower in FID (5.34 against 6.46; $-1.1$, $[-1.4, -0.8]$), OpenDWM has lower FVD-1 (23.8 against 26.9; $+3.1$, $[0.1, 6.0]$), and the FVD-2 difference is not significant ($+2.9$, $[-2.9, 8.7]$).
Its flicker ratio is 1.23 ($[1.15, 1.32]$).
Its object recall is 64.8 against PanDWM's 62.6 ($-2.2$, $[-4.3, -0.1]$), and the road IoUs do not differ significantly (75.7 against 75.5; $-0.2$, $[-0.9, 0.5]$).

\paragraph{Comparison with DiST-4D}
DiST-4D receives one recorded frame, whereas PanDWM receives three.
In the band, DiST-4D reaches FID 17.2 ($[16.0, 18.5]$), FVD-1 55.8 ($[50.8, 60.8]$) and FVD-2 71.0 ($[63.3, 78.7]$).
PanDWM is lower in all three, with differences of $-7.1$ ($[-8.2, -6.1]$), $-18.4$ ($[-23.3, -13.6]$) and $-8.9$ ($[-16.3, -1.5]$), respectively.
In the six cameras, PanDWM is lower in FID ($-6.6$, $[-7.2, -6.0]$) and FVD-1 ($-8.9$, $[-12.1, -5.8]$), and the FVD-2 difference is not significant ($+3.0$, $[-1.7, 7.7]$).
The detector recovers fewer annotated objects in DiST-4D's views (recall 41.4 against PanDWM's 62.6), mostly pedestrians (13.5 against 39.4).
DiST-4D's road IoU is lower than PanDWM's by 3.2 (difference $-3.2$, $[-4.0, -2.5]$).

\paragraph{Comparison with OmniNWM}
OmniNWM receives one recorded frame and the ego trajectory, but no boxes or map.
In the band, it reaches FID 25.4 ($[23.2, 27.7]$), FVD-1 73.0 ($[66.3, 79.8]$) and FVD-2 113.9 ($[102.5, 125.3]$); PanDWM is lower in all three, with differences of $-15.3$ ($[-17.6, -13.1]$), $-35.7$ ($[-42.1, -29.3]$) and $-51.8$ ($[-62.5, -41.1]$).
In the six cameras, PanDWM is lower in FID ($-9.0$, $[-9.9, -8.2]$), FVD-1 ($-16.3$, $[-20.6, -11.9]$) and FVD-2 ($-28.8$, $[-37.1, -20.4]$).
Its flicker ratio of 0.80 ($[0.61, 1.00]$) comes with a sharpness ratio of 0.52.
Without boxes, it generates fewer of the annotated objects (recall 30.4 against PanDWM's 62.6), although the vehicles it does generate are placed well (matched-box IoU 0.69, against 0.73 for the recorded views); PanDWM's road IoU is 4.2 higher ($[3.0, 5.3]$).

\subsection{Test-S statistics}
\label{sec:suppl:tests_stats}
On Test-S, PanDWM reaches FID $7.85\pm0.07$ (95\% interval $[7.2, 8.5]$), FVD-1 $27.3\pm0.5$ ($[25.2, 29.3]$) and FVD-2 $48.5\pm1.0$ ($[44.1, 52.9]$) over three seeds.
These numbers come from different clips and a different sample size than the Val-150 comparison in \cref{tab:main} and are not directly comparable with it.
At half the sample size, two scene-disjoint halves of the recorded clips score FID 34.4, FVD-1 115.5 and FVD-2 118.6, and our generated clips score 1.1, 11.8 and 23.8 above these reference distances.
Relative to these references, the remaining gap is larger in FVD than in FID, and larger in the second window, which is conditioned on generated frames, than in the first.

\subsection{Self-generated history over long rollouts}
\label{sec:suppl:selfhist}
In the ablation of \cref{sec:exp:ablation}, recorded instead of self-generated history lowers FVD-1 by 2.8, but those clips contain only two generated windows.
We therefore roll out both 500-iteration students, the full recipe and the recorded-history variant, over the 12 windows of Test-L with the protocol of \cref{sec:exp:long} (83 scenes, seeds 42 and 0).
The variant has lower FVD only in the first window ($-3.1$); from window~6 on, its FVD is significantly higher in every window.
Over windows 9--11, its FVD is 254.8 against 228.8 for the full recipe ($+26.1$, $[12.6, 39.6]$), and its FVD rises by 21.6 more from windows 3--5 to windows 9--11 ($[6.4, 36.7]$); over windows 3--5, the difference is not significant ($+4.5$, $[-1.1, 10.1]$).
In FID, the variant is higher by 1.2 over windows 3--5 ($[0.1, 2.3]$), and the difference over windows 9--11 is not significant ($+2.4$, $[-1.1, 5.9]$).
Both students are better than the starting student in all three window groups; the starting student reaches FVD 304.5 over windows 9--11 (seed 42).
These 500-iteration students are weaker than our final model.
Intervals of the window groups are delete-one-scene jackknife intervals; per-window significance uses a paired scene bootstrap with 1000 draws.

\subsection{Long-horizon comparison with baselines}
\label{sec:suppl:longbase}
We roll out UniMLVG and OpenDWM (SD3.5) over the 19.5\,s of Test-L with the configurations of \cref{tab:main}, extending only the clip length to 195 frames.
Each continues from its own last three generated frames every 16 frames, the same history length and stride as ours, with the same boxes, maps and camera poses.
Their six views are stitched into the band as in \cref{tab:main} and scored with the instrument of \cref{sec:exp:long} (seed 42); our and the teacher's values are those reported there (\cref{tab:suppl:longbase}).

\begin{table}[t]
  \caption{Long-horizon comparison on Test-L (83 scenes, seed 42), pooled over window groups. Recorded views are the $256\times448$ camera views stitched in the same way. Best method values are bold; $^\dagger$references, not considered.}
  \label{tab:suppl:longbase}
  \centering
  \small
  \setlength{\tabcolsep}{2pt}
  \begin{tabular}{@{}lcccccc@{}}
    \toprule
    & \multicolumn{3}{c}{FID$\downarrow$} & \multicolumn{3}{c}{FVD$\downarrow$} \\
    \cmidrule(lr){2-4}\cmidrule(l){5-7}
    Method & 3--5 & 6--8 & 9--11 & 3--5 & 6--8 & 9--11 \\
    \midrule
    UniMLVG~\cite{unimlvg} & 37.7 & 46.7 & 51.8 & \textbf{75.8} & \textbf{88.6} & \textbf{101.4} \\
    OpenDWM (SD3.5)~\cite{opendwm} & 50.7 & 58.2 & 61.4 & 136.9 & 141.8 & 155.6 \\
    PanDWM (ours) & \textbf{26.1} & \textbf{38.9} & \textbf{49.0} & 86.6 & 113.4 & 126.6 \\
    \midrule
    Causal teacher$^\dagger$ & 27.8 & 43.0 & 45.9 & 116.7 & 136.9 & 157.9 \\
    Recorded views$^\dagger$ & 17.6 & 17.5 & 17.3 & 23.0 & 23.3 & 23.7 \\
    \bottomrule
  \end{tabular}
\end{table}

UniMLVG has lower FVD than PanDWM in every window group; the difference is not significant over windows 3--5 ($-10.8$, $[-23.1, 1.5]$) and significant over windows 6--8 ($-24.8$, $[-39.8, -9.7]$) and 9--11 ($-25.2$, $[-42.0, -8.5]$).
Its FVD rises by 25.7 from windows 3--5 to windows 9--11, against 40.1 for PanDWM, whose FVD is lower only in the first window (58.0 against 71.3).
Our FID is lower over windows 3--5 ($-11.6$, $[-15.3, -7.9]$) and 6--8 ($-7.7$, $[-13.0, -2.5]$); over windows 9--11, the difference is not significant ($-2.8$, $[-10.4, 4.7]$).
Part of the FID difference reflects resolution: the recorded views at $256\times448$, stitched in the same way, score FID 17.3 to 17.6.
OpenDWM has higher FID and FVD than PanDWM in every group; its FVD is higher by 50.3, 28.4 and 29.0, all significant.
Intervals are delete-one-scene jackknife intervals of the differences.
On one A800, generation takes 10.9\,s per frame for UniMLVG and 12.0\,s for OpenDWM.

\subsection{Inference cost at matched angular resolution}
\label{sec:suppl:matchres}
The panorama has 11.4 and 7.7 pixels per degree horizontally and vertically, against about 7 and 6.5 for front and side camera views at $256\times448$, so it carries 1.71 times as many tokens per frame (4608 against 2688).
To separate the representation from this resolution, we time the same four-step student on the two halves and on six camera views at three further sizes (\cref{tab:suppl:matchres}): $304\times736$ matches the panorama's average angular resolution, $320\times832$ makes the views nowhere coarser than the panorama, and $448\times800$ is the output size of DreamForge, DiST-4D and OmniNWM.
All six views share one size, so the wider rear camera stays coarser than the panorama in every configuration.
The resized views are upsampled from $256\times448$, so only their cost is measured; all runs use one H100, the same weights, clips and seeds, and the time scope of \cref{tab:main}, denoising plus decoding.
At the average angular resolution, six views cost as much as the two halves (0.479 against 0.495\,s per frame).
Six views that are nowhere coarser need 6240 tokens and take 1.17 times as long, and six $448\times800$ views take 1.66 times as long: camera images sample their periphery more densely and must cover extra elevation at their centres to reach $\pm16.7^\circ$ at their edges.

\begin{table}[t]
  \centering
  \caption{Denoising and decoding time of the same four-step student on one H100, in seconds per generated frame (mean of three clips, text encoder on the GPU). Relative: time divided by that of the two halves.}
  \label{tab:suppl:matchres}
  \resizebox{\linewidth}{!}{
  \begin{tabular}{@{}llrrrrr@{}}
    \toprule
    Input & Size & Tokens & Denoise & Decode & Total & Relative \\
    \midrule
    Two halves & $2\times256\times2304$ & 4608 & 0.343 & 0.152 & 0.495 & 1.00 \\
    Six views & $6\times256\times448$ & 2688 & 0.169 & 0.072 & 0.241 & 0.49 \\
    Six views, average resolution & $6\times304\times736$ & 5244 & 0.339 & 0.141 & 0.479 & 0.97 \\
    Six views, nowhere coarser & $6\times320\times832$ & 6240 & 0.410 & 0.170 & 0.579 & 1.17 \\
    Six views & $6\times448\times800$ & 8400 & 0.573 & 0.250 & 0.824 & 1.66 \\
    \bottomrule
  \end{tabular}}
\end{table}

\section{Reproduction of baselines}
\label{sec:suppl:baselines}
\paragraph{Methods without a reproduction}
Panacea~\cite{panacea}, DriveScape~\cite{drivescape} and WorldSplat~\cite{worldsplat} could not be run under our protocol: DriveScape and WorldSplat have released no nuScenes weights, and Panacea only those of its later Panacea+ video stage.
Their reported nuScenes validation numbers use different protocols and are not comparable with our tables.
Panacea reports FID 16.96 and FVD 139 for 8-frame videos at $256\times512$ per view with a generated first frame.
DriveScape reports FID 8.34 and FVD 76.39 for 8-frame videos at $576\times1024$, conditioned on the first multi-view frames and resized to $224\times400$ for evaluation.
WorldSplat reports FID 8.78 and FVD 74.13 without and 4.14 and 16.57 with a first-frame condition.

\paragraph{UniMLVG}
We run the released checkpoint with the OpenDWM code (commit b0ecc3d) and its stage-3 inference configuration: windows of 19 frames with 3 reference frames, 50 steps, guidance 3 and six views at $256\times448$.
The VAE and text encoders are loaded from SD3.5-medium, which our model also uses, in place of the SD3-medium named in its configuration.
Its validation clips are the scene-start clips of the 150 validation scenes; for every clip, the first LiDAR timestamp equals the anchor of the Val-150 clip.
In the protocol of the evaluation configuration of UniMLVG~\cite{unimlvg} released in the OpenDWM repository~\cite{opendwm} (per view at $256\times448$, FID over the 16 frames after the reference frames of all 900 view videos, FVD of the TATS I3D network on the same 16 frames), our runs reach FID 5.83 and FVD 35.7 with 3 recorded frames, against 5.8 and 36.1 reported in~\cite{unimlvg}, and FVD 101.3 without recorded frames, against 106.01 listed in that configuration.

\paragraph{OpenDWM (SD3.5)}
We run the released ctsd\_35\_tirda\_bm\_nwao checkpoint (40k steps, trained on nuScenes, Waymo, Argoverse 2 and OpenDV) with the same code and the inference settings of its configuration: windows of 19 frames with 3 reference frames, 40 steps, guidance 4 and six views at $256\times448$.
Its validation clips are those of UniMLVG.
In the protocol of UniMLVG, our run reaches FVD 40.3 with 3 recorded frames, against 38.83 listed in its configuration file.

\paragraph{MagicDrive-V2}
We run the released stage-3 checkpoint (moving average) with the official code (commit 4ed72c6) and its $424\times800$ inference configuration: 30 rectified-flow steps and guidance 2.
Clip $i$ of its validation set starts at the first frame of validation scene $i$, so the clips are again the scene-start clips.
We generate 49 frames at 12\,Hz and take, for each of our 10\,Hz frames, the 12\,Hz frame with the same LiDAR timestamp.
The $424\times800$ output omits the top 26 rows of the half-resolution view; we fill them by repeating the first row before stitching, and exclude them from the six-camera mask for all methods.
MagicDrive-V2 reports its results at $848\times1600$; we did not reproduce them.

\paragraph{DiST-4D}
We run the released DiST-T checkpoint with the official code (commit 1a33521) and its $424\times800$ inference configuration: 30 rectified-flow steps and guidance 2.
The model takes exactly one recorded reference frame, so it receives the latest of our three history frames.
As for MagicDrive-V2, we generate 49 frames at 12\,Hz, take for each of our 10\,Hz frames the 12\,Hz frame with the same LiDAR timestamp, and fill and mask the 26 missing top rows.
In its setting (17 frames at 12\,Hz, one reference frame, FID and FVD per view at $424\times800$ over the 16 frames after the reference), with one clip from the first frame of each of the 150 validation scenes, our run reaches FID 13.7 and FVD 57.1, against 6.83 and 22.67 reported in~\cite{dist4d}.
The paper does not state the number of clips behind these values, and we could not close the gap; we report the numbers as measured.

\paragraph{OmniNWM}
We run the released checkpoint with the official code (commit 8e2333e) and its $448\times800$ inference configuration on eight GPUs with its sequence parallelism: 50 steps, whose sampler has no guidance branch.
The model takes one reference frame, for which we use the latest of our three history frames, and the recorded ego trajectory as ray maps; it receives no boxes, map or text.
A window has 33 frames at 12\,Hz, so we roll out a second window from the last generated frame, as in its long-rollout mode, and give it the recorded trajectory of that window; the released multi-round loop would reuse the trajectory of the first window.
As for MagicDrive-V2, we take for each of our 10\,Hz frames the 12\,Hz frame with the same LiDAR timestamp, and the 448 output rows are padded by repeating the first and last rows.
The repository contains no FID or FVD code, so we could not check its reported values.

\paragraph{DreamForge}
We run the released DreamForge-DiT temporal checkpoint with the official code (DriveArena, commit c71a020) and its $448\times800$ inference configuration: 30 rectified-flow steps and guidance 2.
Each window has 25 frames at 12\,Hz, the first 9 of which are past frames.
In the first window, our three history frames fill these 9 slots as the dataset fills them at scene starts (the first frame seven times, then the other two); each later window takes the last 9 generated frames.
Three windows give 48 new frames, which we map to our 10\,Hz frames by LiDAR timestamp as for MagicDrive-V2; the 2 rows missing at the top of each view are filled in the same way.
To check the setup, we run the official validation script on the 150 validation scenes (17 frames at 12\,Hz from the scene start, without recorded frames) and score the 16-frame video it writes for each view with the FVD code of the W-CODA 2024 challenge, track 2 (I3D, $224\times400$).
This gives FVD 80.6, against 103.61 reported in~\cite{dreamforge} for its DiT variant on 16-frame clips at $448\times800$; the paper does not name its clips or FVD code, so the two values are not strictly comparable.

\paragraph{RAYNOVA}
We run the released checkpoint, trained on public nuScenes and nuPlan data, with the official code (commit 8546867) through its interactive inference path at $384\times672$, conditioned on our three history frames, projected boxes, map points and text, with 0.1\,s between frames (its released configuration samples frame gaps of up to 0.5\,s).
Its nuScenes converter is not released, so we convert the clips to its ScenarioNet format ourselves, and its scene captions are replaced by those given to UniMLVG.
In the band, it reaches FID 34.0, FVD-1 248.6 and FVD-2 264.2 (six cameras: 17.1, 177.0 and 197.5), and its flicker ratio is 2.97: single frames look realistic, but static content changes from frame to frame.
With the nuScenes defaults of its inference notebook (2\,Hz keyframes, no recorded frames, 16 frames), we obtain FID 16.7 and FVD 213.6 per view, against 10.5 and 91 reported in~\cite{raynova}, whose evaluation protocol is not stated.
Its nuScenes input relies on our own conversion, with which it neither reproduces its reported numbers nor keeps static content stable, so we report it here rather than in \cref{tab:main}.

\paragraph{Generation time}
Generation time is measured on one A40 for Val-150 clips 1 to 3, after one warm-up clip (OmniNWM on three windows of clips 0 and 1 after one warm-up window), and counts denoising and the decoding of the RGB latents.
DiST-4D denoises RGB and depth latents jointly in one 49-frame window with batched guidance (60 network evaluations) and takes 13.2\,s per new 12\,Hz frame; decoding its depth latents, which the scored views do not use, would add 0.8\,s.
DreamForge evaluates the two guidance branches in separate calls (60 evaluations per window) and takes 29.9\,s per new 12\,Hz frame.
Per 0.1\,s of video, as in the main table, these are 15.9\,s and 35.9\,s.
OpenDWM (SD3.5) batches its two guidance branches (80 evaluations per window) and takes 22.9\,s per 10\,Hz frame.
OmniNWM runs 50 network evaluations per window; on one A40 without sequence parallelism it takes 74.9\,s per new 12\,Hz frame, i.e. 89.9\,s per 0.1\,s of video, and 7.7\,s per 0.1\,s on eight A800 GPUs.

\paragraph{Gaussian reconstruction baselines}
All Gaussian baselines use their released weights at their own resolution and are rendered into the target cameras with the same intrinsics convention (MCSplat with its own, below); for every adapter, a single Gaussian at a known 3D point lands within 0.015\,px of its projection.
Omni-Scene takes the six images of the input frame with their Metric3D-v2~\cite{metric3dv2} depth and confidence, DrivingForward runs its single-frame model, ReconDrive takes two frames and MCSplat one frame with a LiDAR sweep; our 3D Gaussians use the input frame, and our 4D Gaussians add the four recorded frames 0.5--2\,s before it.
ReconDrive (commit d2bc397, stage-2 weights) receives the input frame and the frame six interpolated 12\,Hz samples earlier (0.50\,s on average), and vehicles move with the velocities of their ground-truth boxes, as in its released inference configuration; without these velocities, PSNR changes by 0.03\,dB.
It is trained to render views between its two inputs, so it extrapolates to the targets after the input frame.
On 2 of its 14 evaluation scenes, its own evaluation gives reconstruction PSNR 32.38 and novel-view PSNR 21.99, against 32.66 and 23.99 reported in~\cite{recondrive} as means over all 14 scenes, so the values are not directly comparable.
The released model gives the Gaussians of the later input frame much lower opacity than those of the earlier one (0.22 against 0.81 on average over seven samples), which is consistent with its highest score on the target 3.2\,m behind.

MCSplat (commit 9b774b2, nuScenes weights) takes one frame of six cameras at $406\times224$, sky masks and a LiDAR sweep, whose projected depth is used to match colours across cameras; we compute the sky masks with MoGe-2~\cite{moge2}, which reproduces the masks of its released example scene (pixel agreement above 0.9999).
We render with its own intrinsics convention, which is 0.6\,dB higher on its input views than adding the half-pixel offset of the other adapters; its renders are partly transparent (mean opacity 0.94), so the gray fill of uncovered pixels lowers its PSNR.
Its code release has no evaluation script; following its protocol on the 46 example bins released with Omni-Scene, our evaluation gives PSNR 24.16, against 26.19 reported in~\cite{mcsplat} for the full validation set (Omni-Scene: 25.48 on the same bins).

\paragraph{Two-stage baseline for generated scenes}
The two-stage row of \cref{tab:gs_gen} reconstructs the DiST-4D videos of the DiST-4D row with ReconDrive.
ReconDrive receives two generated frames 0.5\,s apart (clip frames $c$ and $c+5$ for $c=3,8,\dots,28$, and 29 and 34 for the last two frames) in place of recorded images, with the boxes and velocities of its released configuration, and each frame is rendered from the pair that starts at or just before it.
We render at its native $280\times518$ resolution, because its pixel-aligned Gaussians leave gaps at $450\times800$ (coverage 0.86 against 1.00 on the first frame of one clip); the views are then stitched into the band, and uncovered pixels are filled as for our Gaussians.
On the same pair of that clip, its own reconstruction of the earlier input frame reaches 29.9\,dB with generated input and 32.3\,dB with the recorded frames.

\paragraph{Front-view world models}
We run the public checkpoints of HorizonDrive~\cite{horizondrive}, Epona~\cite{epona}, Vista~\cite{vista} and GEM~\cite{gem} on the scene-start clips of 149 validation scenes with seed 42.
We compare their front-view outputs with the front view resampled from our panoramas in \cref{fig:frontview}.
Raw camera images were unavailable on the cluster for these runs, so each input frame is resampled from the recorded panoramic band using an ideal spherical model.
The inputs cover only the rows represented by the band and are not pixel-aligned with the recorded camera images.
HorizonDrive receives the first three resampled front frames instead of its default eleven, together with ego actions, 3D boxes and HD maps, and samples with four steps.
Epona receives ten resampled frames and the recorded ego trajectory and samples with 100 steps per frame.
Vista and GEM receive one resampled frame and sample with 50 steps.
The differing inputs make this a qualitative comparison rather than a ranking.

The four illustrated clips are evenly spaced in time among the 149 scene-start clips and were fixed before our outputs were viewed.
The figure shows predictions 2.5\,s after each clip starts.
Vista and GEM produce 16:9 frames with the same field of view, which we resize for display.
Dashed ellipses identify corresponding objects.
In the second row, HorizonDrive turns part of the articulated bus into a truck, Epona and GEM blur it, and Vista leaves it near its earlier position.
In the third row, a white vehicle enters at the left edge; HorizonDrive (unmarked) and PanDWM generate it.

\paragraph{Multi-camera comparison}
\Cref{fig:multiview} shows PanDWM, UniMLVG and OmniNWM in panorama space; the front-view models appear in \cref{fig:frontview}.
OmniNWM is run as described in \cref{sec:suppl:baselines}, and the views of both baselines are stitched with the OneBEV mapping of \cref{tab:main}.
The three clips were fixed before any output was viewed: four clips were spaced evenly among the 15 clips of an earlier OmniNWM run, and the night clip was then excluded.
The images show the front-left, front and front-right camera-owned sectors at 2.5\,s, with one row per method and about $\pm13^\circ$ elevation.
Ground truth and PanDWM use their native panoramas, and all methods use seed 42.
Brightness steps at sector boundaries come from the separate camera exposures.
In the third clip, a truck passes on the left; OmniNWM generates a cab in the front-left sector where the recording shows the trailer.

\paragraph{Novel views in the first figure}
The novel views of \cref{fig:teaser} render the Gaussians of frame 18 of one generated clip (nuScenes scene-0096) with a $90^\circ$ horizontal field of view, using the decoder and the Depth Anything 3~\cite{da3} relabelling of the generated-scene evaluation (Sec.~\ref{sec:suppl:gseval}).
As for the 4D Gaussians of \cref{sec:method:gs}, the static Gaussians of frames $t-20$, $t-15$, $t-10$, $t-5$, $t+5$ and $t+10$ fill the pixels that frame $t$ does not cover, and dynamic content comes only from frame $t$.
Unlike the evaluation, which removes frame $t$'s Gaussians at depth discontinuities and those larger than 1\,m, the figure keeps all of them, since the removal outlines objects with holes in views without translation.
The pixels that then remain uncovered (0.6\%, 3.4\% and 0.04\% in the reference, forward and turned views) are filled by push-pull interpolation, which the generated-scene evaluation applies in the same way to the uncovered pixels of every method.

\begin{table*}[t]
  \caption{Setting G (conditions only) on Val-150 in the two evaluation spaces of the main paper. PanDWM: mean over three seeds; baselines: seed 42. NFE counts denoiser evaluations per window with both guidance branches; the window is 19 frames for UniMLVG and PanDWM and the whole 49-frame clip for MagicDrive-V2.}
  \label{tab:suppl:settingG}
  \centering
  \small
  \setlength{\tabcolsep}{4pt}
  \begin{tabular}{@{}lccccccccc@{}}
    \toprule
    & & & & \multicolumn{3}{c}{Panoramic band} & \multicolumn{3}{c}{Six cameras, $256\times448$} \\
    \cmidrule(lr){5-7}\cmidrule(lr){8-10}
    Method & Views & Hist. & NFE & FID$\downarrow$ & FVD-1$\downarrow$ & FVD-2$\downarrow$ & FID$\downarrow$ & FVD-1$\downarrow$ & FVD-2$\downarrow$ \\
    \midrule
    MagicDrive-V2~\cite{magicdrivev2} & 6 & 0 & 60 & 45.9 & 141.3 & 134.1 & 21.93 & 86.0 & 78.6 \\
    UniMLVG~\cite{unimlvg} & 6 & 0 & 100 & 42.6 & 109.7 & 104.2 & 16.03 & 73.7 & 74.6 \\
    PanDWM (ours) & 2 & 0 & 4 & 49.3 & 242.1 & 206.0 & 27.25 & 195.7 & 171.4 \\
    \bottomrule
  \end{tabular}
\end{table*}

\begin{table*}[t]
  \caption{Seam ratios at the six camera boundaries of the stitched band (pooled) on Val-150, and the excess over recorded views at the same resolution, stitched in the same way (95\% intervals). Rear joint: the column where the two halves of PanDWM meet without overlap; the recorded band reads 0.85 there.}
  \label{tab:suppl:seams}
  \centering
  \small
  \setlength{\tabcolsep}{3pt}
  \begin{tabular}{@{}lccc@{}}
    \toprule
    Method & Ratio & Excess at cameras & Excess at rear joint \\
    \midrule
    UniMLVG~\cite{unimlvg} & 14.71 & $-0.24$ $[-0.42, -0.08]$ & -- \\
    OpenDWM (SD3.5)~\cite{opendwm} & 16.27 & $+1.32$ $[1.10, 1.56]$ & -- \\
    DreamForge~\cite{dreamforge} & 12.70 & $+2.09$ $[1.82, 2.37]$ & -- \\
    DiST-4D~\cite{dist4d} & 12.19 & $+1.57$ $[1.20, 1.92]$ & -- \\
    OmniNWM~\cite{omninwm} & 12.97 & $+2.35$ $[2.11, 2.60]$ & -- \\
    \midrule
    Causal teacher & 10.22 & $-0.07$ $[-0.18, 0.05]$ & $+1.99$ $[1.85, 2.13]$ \\
    PanDWM (ours) & 9.99 & $-0.30$ $[-0.42, -0.17]$ & $+2.15$ $[1.99, 2.31]$ \\
    \bottomrule
  \end{tabular}
\end{table*}

\section{Setting G}
\label{sec:suppl:settingG}
In setting G, a method receives only the conditions (3D boxes, road map, camera poses and text) and generates all 35 frames of the Val-150 clips; frames 3 to 34 are scored as in \cref{sec:exp:setup}.
\Cref{tab:suppl:settingG} lists the results in both evaluation spaces.
Our student was trained only behind recorded or self-generated history.
Before reporting it in setting G, we fixed a rule on Dev-78: the history-free student enters setting G if its FID is at most 1.0 and its FVD-1 at most 5 above those of the consistency-distilled student without history.
With seed 42 it passes by a wide margin, with FID 54.4 against 80.7 and FVD-1 271.5 against 418.1.
On the same clips with recorded history, the same model reaches FID 14.1 and FVD-1 57.1 (seed 42), so PanDWM is primarily a predictive model, and its setting-G numbers are not comparable to those with recorded history.
On Val-150 in setting G, PanDWM reaches FID $49.3\pm0.4$ ($[42.7, 55.9]$), FVD-1 $242.1\pm3.4$ ($[211.8, 272.3]$) and FVD-2 $206.0\pm0.9$ ($[181.3, 230.8]$) over three seeds.
Both six-camera baselines are better than PanDWM in setting G.
In the band, UniMLVG reaches FID 42.6, FVD-1 109.7 and FVD-2 104.2, and MagicDrive-V2 45.9, 141.3 and 134.1. The FVD-1 of PanDWM is higher by 132.4 ($[100.6, 164.1]$) than that of UniMLVG and by 100.7 ($[69.1, 132.3]$) than that of MagicDrive-V2, and its FID differences to both are not significant ($+6.7$, $[-0.4, 13.8]$; $+3.4$, $[-4.2, 11.0]$).
In the six cameras, PanDWM is above UniMLVG by 11.22 in FID ($[7.34, 15.09]$) and 122.0 in FVD-1 ($[96.5, 147.4]$), and above MagicDrive-V2 by 5.32 ($[1.28, 9.37]$) and 109.7 ($[85.1, 134.3]$).
Between the two baselines, UniMLVG has the lower FVD-1 in the band (UniMLVG minus MagicDrive-V2: $-31.7$, $[-47.3, -16.0]$) and in the cameras ($-12.2$, $[-23.5, -1.0]$).

On one A40 GPU, including VAE decoding, PanDWM takes 1.79\,s per frame in setting G, UniMLVG 18.5\,s and MagicDrive-V2 14.3\,s per 12\,Hz frame (17.2\,s per 0.1\,s of video).
On static scenes, the frame-to-frame change relative to the recorded band is 2.29 for PanDWM ($[2.01, 2.58]$), 1.56 for UniMLVG ($[1.36, 1.75]$) and 1.73 for MagicDrive-V2 ($[1.37, 2.10]$).
In controllability (protocol of \cref{sec:exp:setup}), object recall is 66.5 for PanDWM, 69.5 for UniMLVG and 50.0 for MagicDrive-V2, and road IoU 75.5, 71.1 and 67.0; PanDWM minus UniMLVG is $-3.0$, $[-5.4, -0.6]$ in recall and $+4.3$, $[2.3, 6.4]$ in road IoU.

\section{Seams at camera boundaries}
\label{sec:suppl:seams}
To measure seams, we average the absolute horizontal luminance difference between neighbouring columns of the stitched band over rows, the 32 scored frames and the 150 Val-150 clips.
At a column $b$, the seam ratio sums this profile over columns $b-2$ to $b+2$, subtracts four times its mean over the columns 4 to 10 away from $b$ on both sides, and divides by that mean; it is close to 1 without a seam and to the step size for a one-column step.
Recorded panoramas already have seams at the six camera boundaries, from exposure and parallax differences (ratio 10.28), and stitching $256\times448$ camera views adds more (14.95).
We therefore report each method's excess over recorded views at its own resolution, stitched in the same way, with 95\% intervals from 2000 scene-bootstrap resamples.

\Cref{tab:suppl:seams} shows that PanDWM adds no seam at the camera boundaries, like UniMLVG and the teacher, whereas OpenDWM, DiST-4D, DreamForge and OmniNWM add 1.3 to 2.4 over their references.
Most of the difference in raw ratios therefore comes from stitching low-resolution camera views rather than from inconsistency between generated views.
Behind the vehicle, where the two halves meet without overlap, PanDWM adds a seam of $+2.15$; the teacher shows the same seam ($+1.99$), so it comes from the two-half representation rather than from distillation.
At the front joint, the 128-column cross-fade removes the seam ($+0.01$, $[-0.04, 0.06]$, against $+0.62$, $[0.56, 0.69]$ for a hard cut).
The ratio is local and does not detect content mismatches that keep local gradients unchanged.
Varying the cross-fade width (0, 32, 64, 128, 256 or 494 of the 494 overlapping columns) changes only the front joint; the rear joint and the camera boundaries are unchanged.
Relative to the 128 columns we use, a hard cut adds a front seam of $+0.62$ and raises FVD-1 by 0.25 ($[0.11, 0.39]$), and 64 columns still raise FVD-1 by 0.08 ($[0.004, 0.15]$).
Wider blends lower FID slightly ($-0.08$ at 494 columns) by blurring the blend zone, whose gradient energy drops by 6.6\%; FVD-2 does not change with the width.

\onecolumn
\section{Frames of generated videos}
\label{sec:suppl:frames}
This section shows frames of our generated videos next to the recorded ones.
The clips and the shown times were fixed by an index rule before any output was viewed: evenly spaced Val-150 and Test-S clips, and Test-L scenes 75, 101 and 131 and three evenly spaced further Test-L scenes.
All videos are generated by the four-step student with seed 42 from three recorded frames; the recorded frames after them are shown for reference only.
Each panel shows the recorded panorama (top) and PanDWM (below) at the marked time after the first of the three recorded frames, as the full $360^\circ$ band with the cameras, from left to right, back, back-left, front-left, front, front-right and back-right.
Val-150 and Test-S clips have 32 generated frames in two autoregressive windows of 16 and are shown at 0.3, 1.9 and 3.4\,s, the first, middle and last generated frames; the 19.5\,s Test-L rollouts have 192 generated frames in 12 windows of 16 and are shown at 0.3, 5, 10, 15 and 19\,s.

\newcommand{\framepanel}[4]{\par\noindent\makebox[\linewidth]{\sbox0{\includegraphics[width=0.98\linewidth,trim=0 #2 0 #3,clip]{supplementary_figures/#1_full.jpg}}\usebox0\llap{\raisebox{\dimexpr\ht0-1.6ex\relax}{\scriptsize\textcolor{gray}{#4}}\hspace{3pt}}}\par}
\newcommand{\cliptitle}[1]{\par\smallskip\noindent{\small\textbf{#1}}\par\nopagebreak\vspace{2pt}}

\subsection{Val-150 clips}
\cliptitle{Val-150 clip 18}
\framepanel{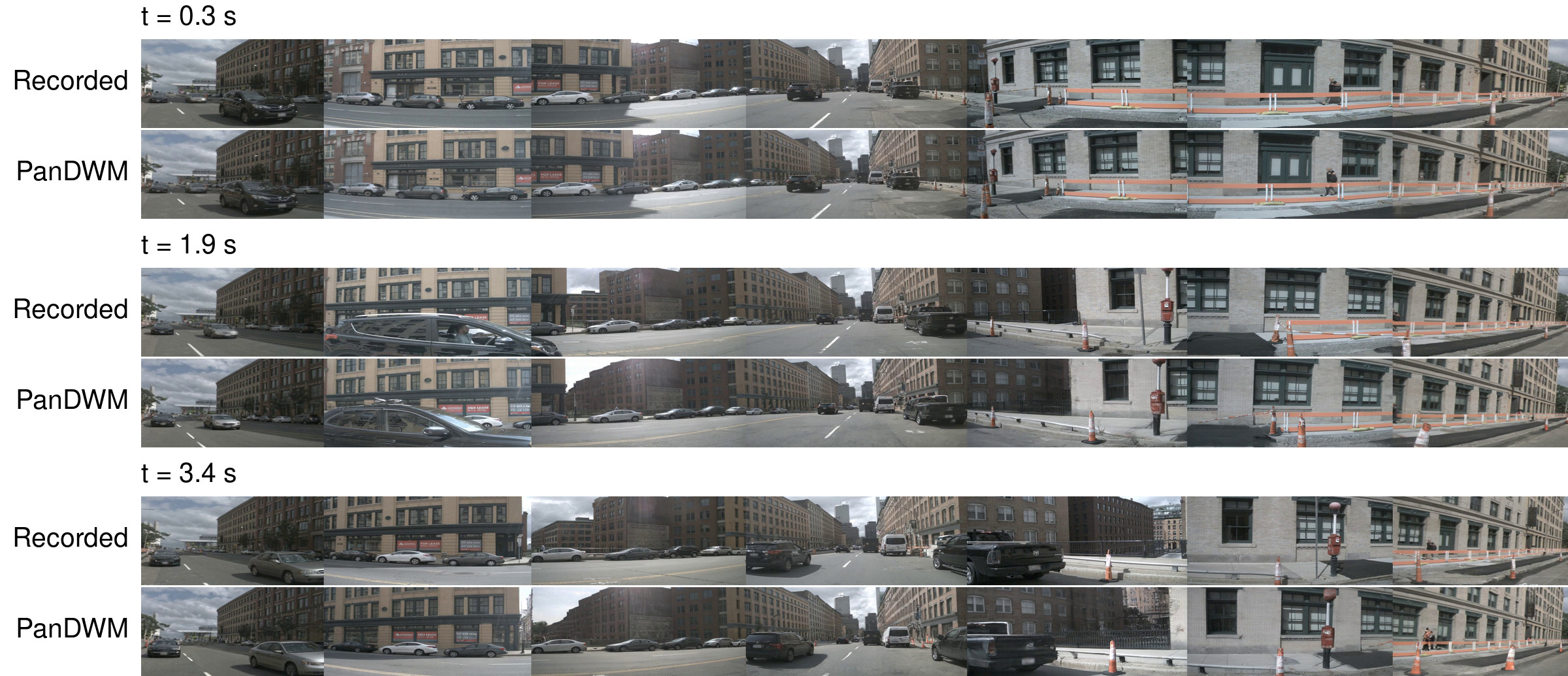}{143.79}{0.00}{Val-150 clip 18}
\framepanel{val150_clip018}{71.59}{72.21}{Val-150 clip 18}
\framepanel{val150_clip018}{0.00}{144.41}{Val-150 clip 18}
\cliptitle{Val-150 clip 56}
\framepanel{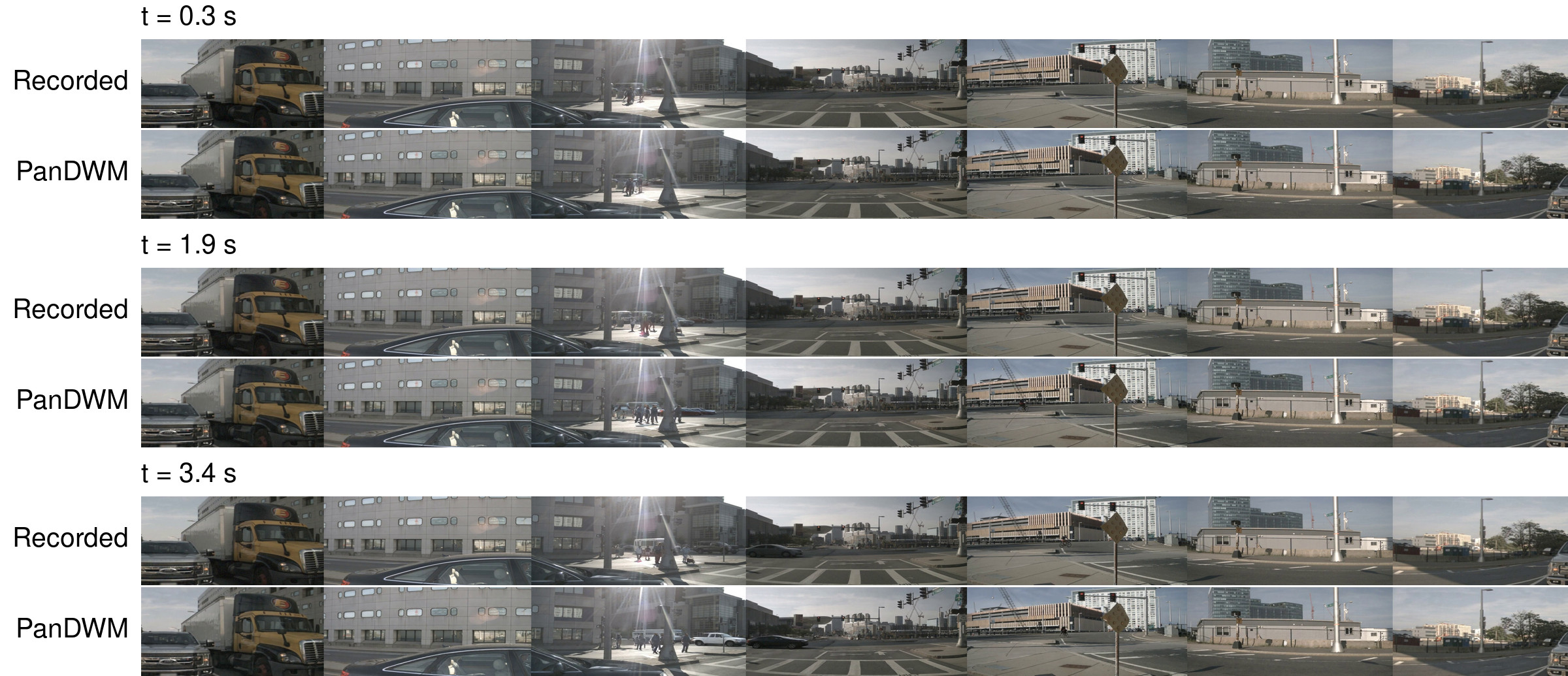}{143.79}{0.00}{Val-150 clip 56}
\framepanel{val150_clip056}{71.59}{72.21}{Val-150 clip 56}
\framepanel{val150_clip056}{0.00}{144.41}{Val-150 clip 56}
\cliptitle{Val-150 clip 93}
\framepanel{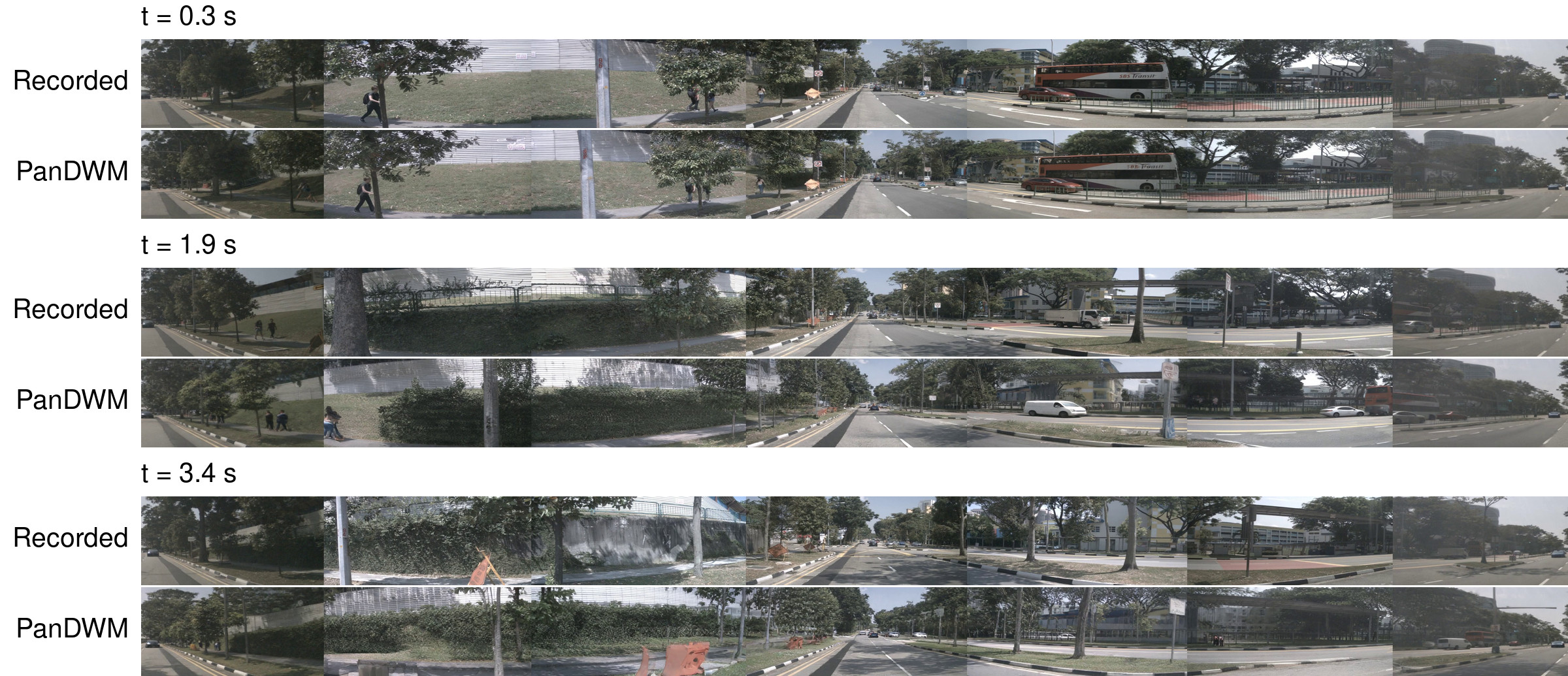}{143.79}{0.00}{Val-150 clip 93}
\framepanel{val150_clip093}{71.59}{72.21}{Val-150 clip 93}
\framepanel{val150_clip093}{0.00}{144.41}{Val-150 clip 93}
\cliptitle{Val-150 clip 131, the same nuScenes scene as Test-L scene 131}
\framepanel{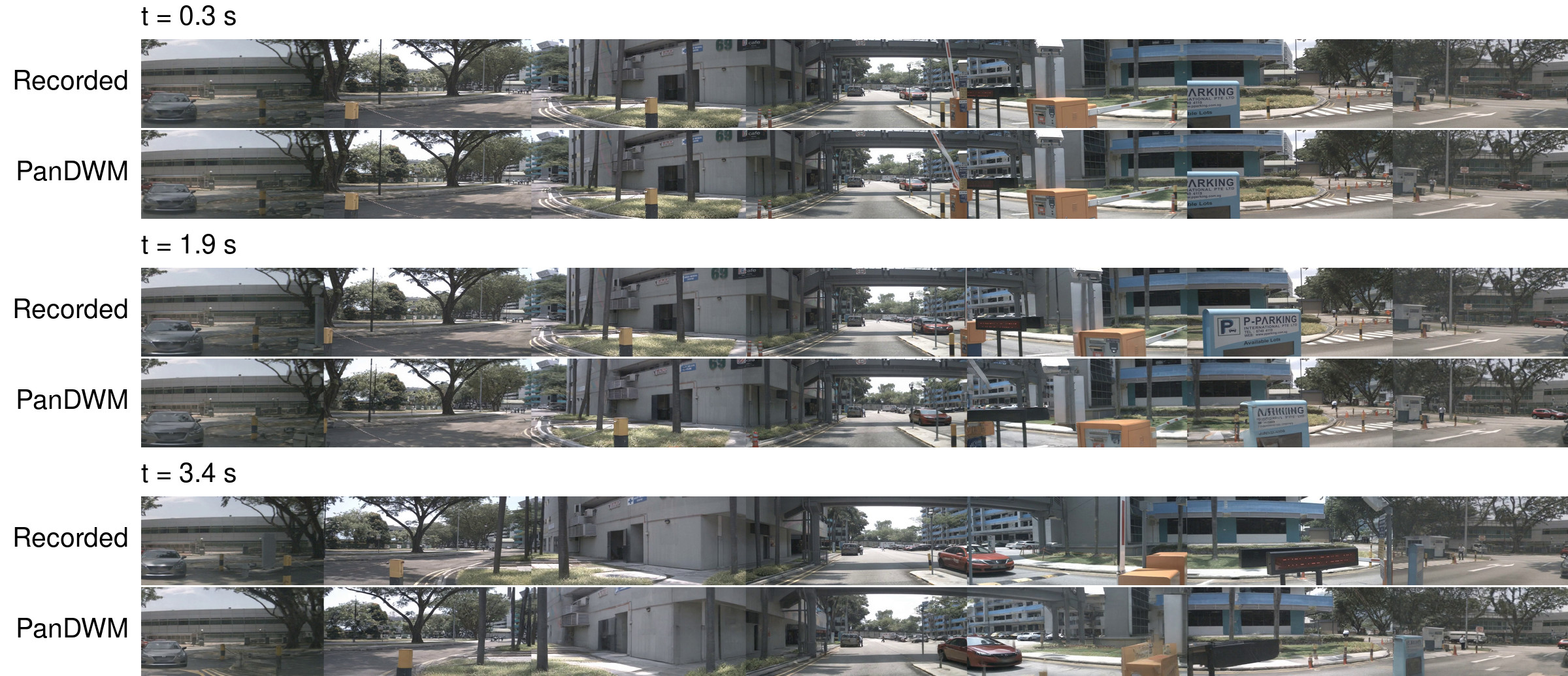}{143.79}{0.00}{Val-150 clip 131}
\framepanel{val150_clip131}{71.59}{72.21}{Val-150 clip 131}
\framepanel{val150_clip131}{0.00}{144.41}{Val-150 clip 131}
\subsection{Test-S clips}
\cliptitle{Test-S clip 31, starting 10\,s into its scene}
\framepanel{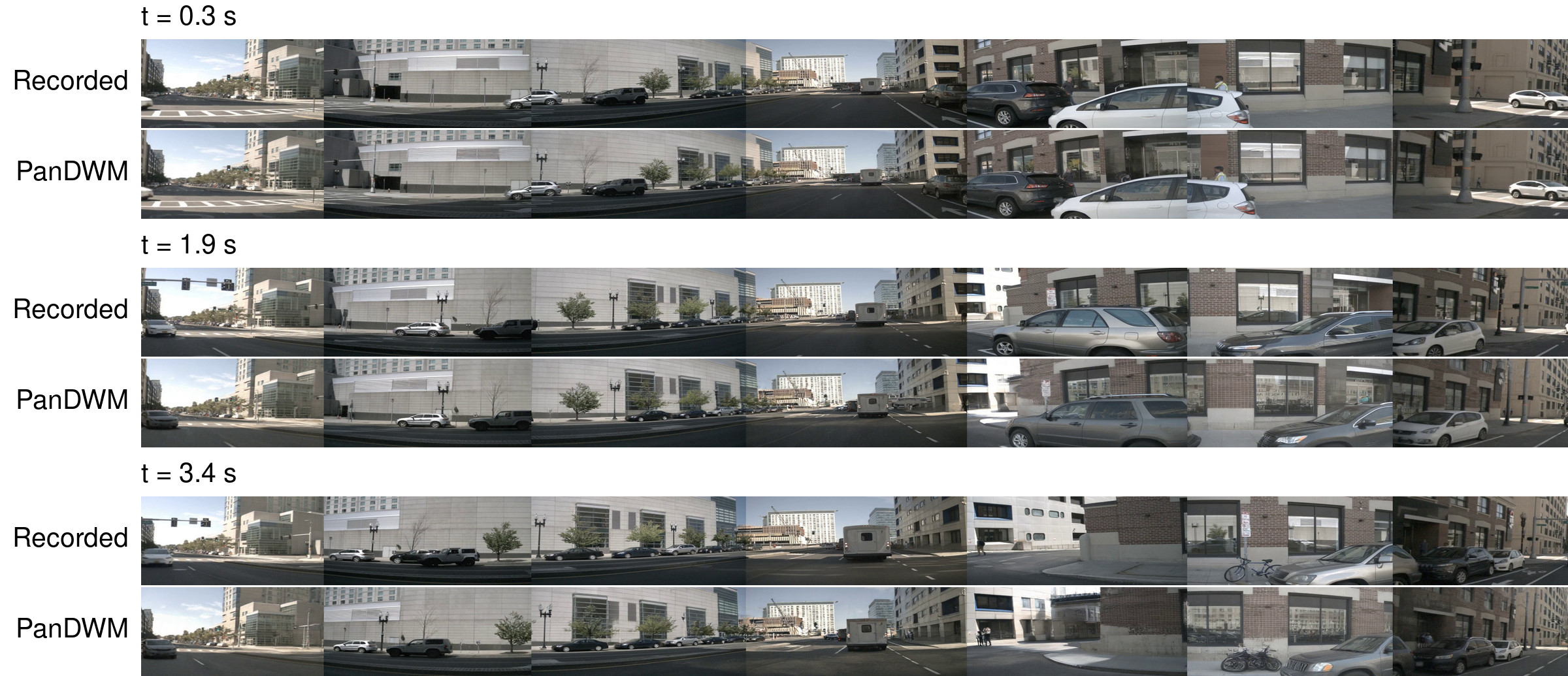}{143.79}{0.00}{Test-S clip 31}
\framepanel{tests_clip031}{71.59}{72.21}{Test-S clip 31}
\framepanel{tests_clip031}{0.00}{144.41}{Test-S clip 31}
\cliptitle{Test-S clip 93, starting 5\,s into its scene}
\framepanel{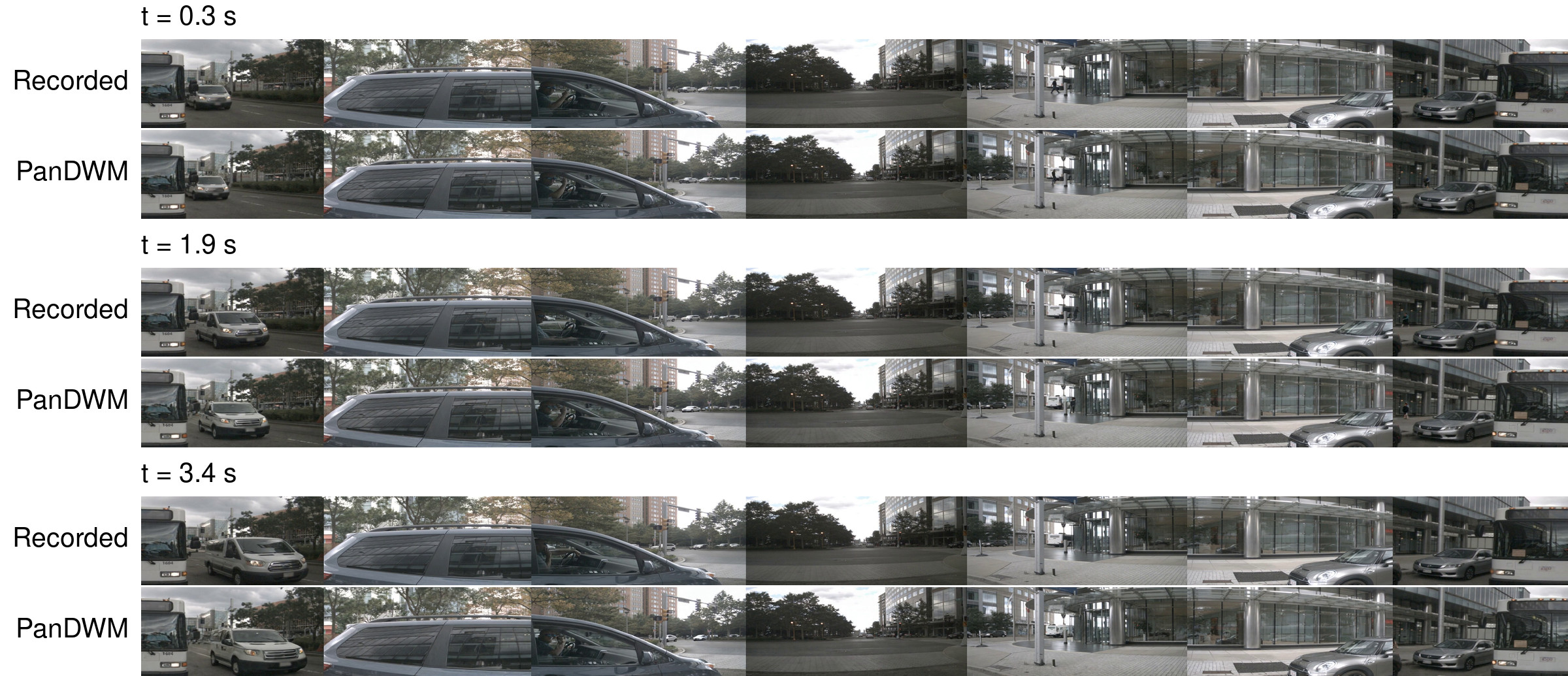}{143.79}{0.00}{Test-S clip 93}
\framepanel{tests_clip093}{71.59}{72.21}{Test-S clip 93}
\framepanel{tests_clip093}{0.00}{144.41}{Test-S clip 93}
\cliptitle{Test-S clip 155, starting 15\,s into its scene}
\framepanel{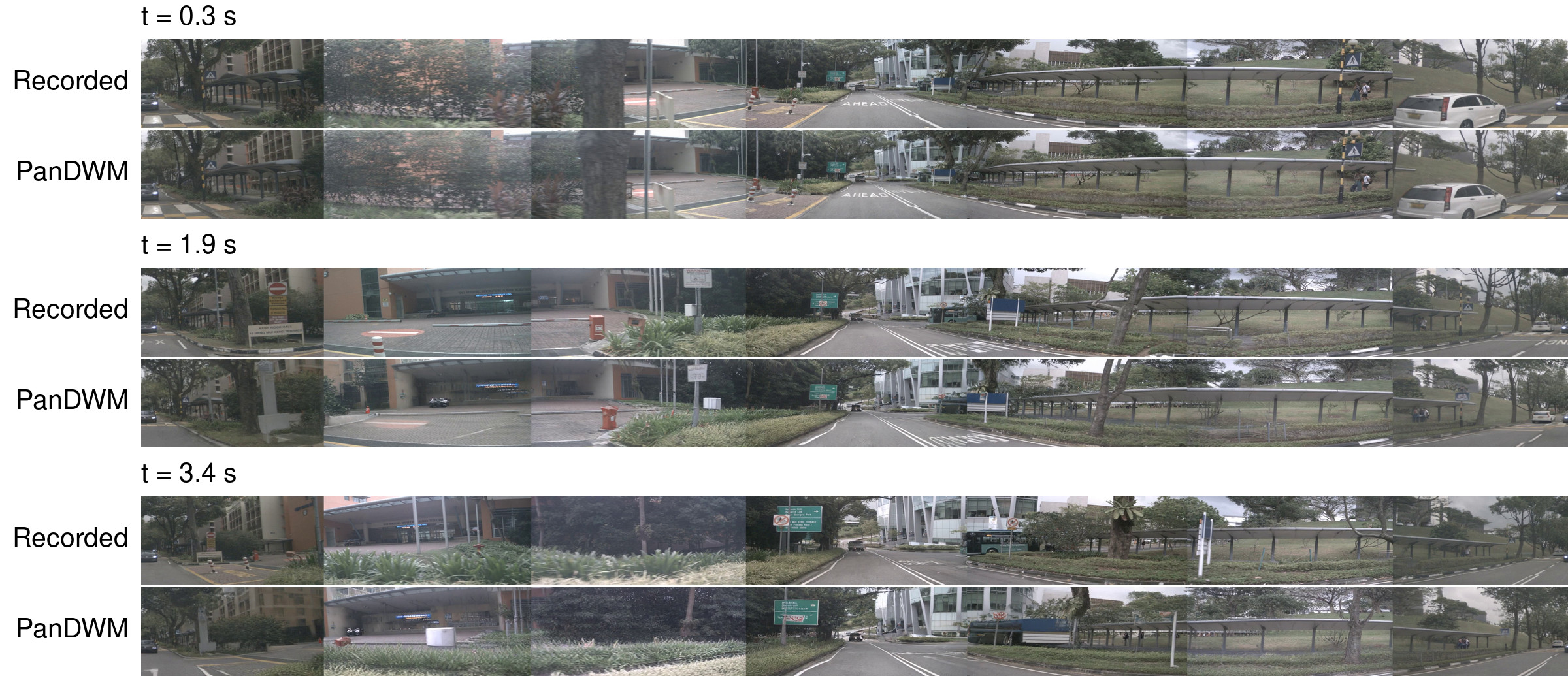}{143.79}{0.00}{Test-S clip 155}
\framepanel{tests_clip155}{71.59}{72.21}{Test-S clip 155}
\framepanel{tests_clip155}{0.00}{144.41}{Test-S clip 155}
\cliptitle{Test-S clip 217, starting 10\,s into its scene}
\framepanel{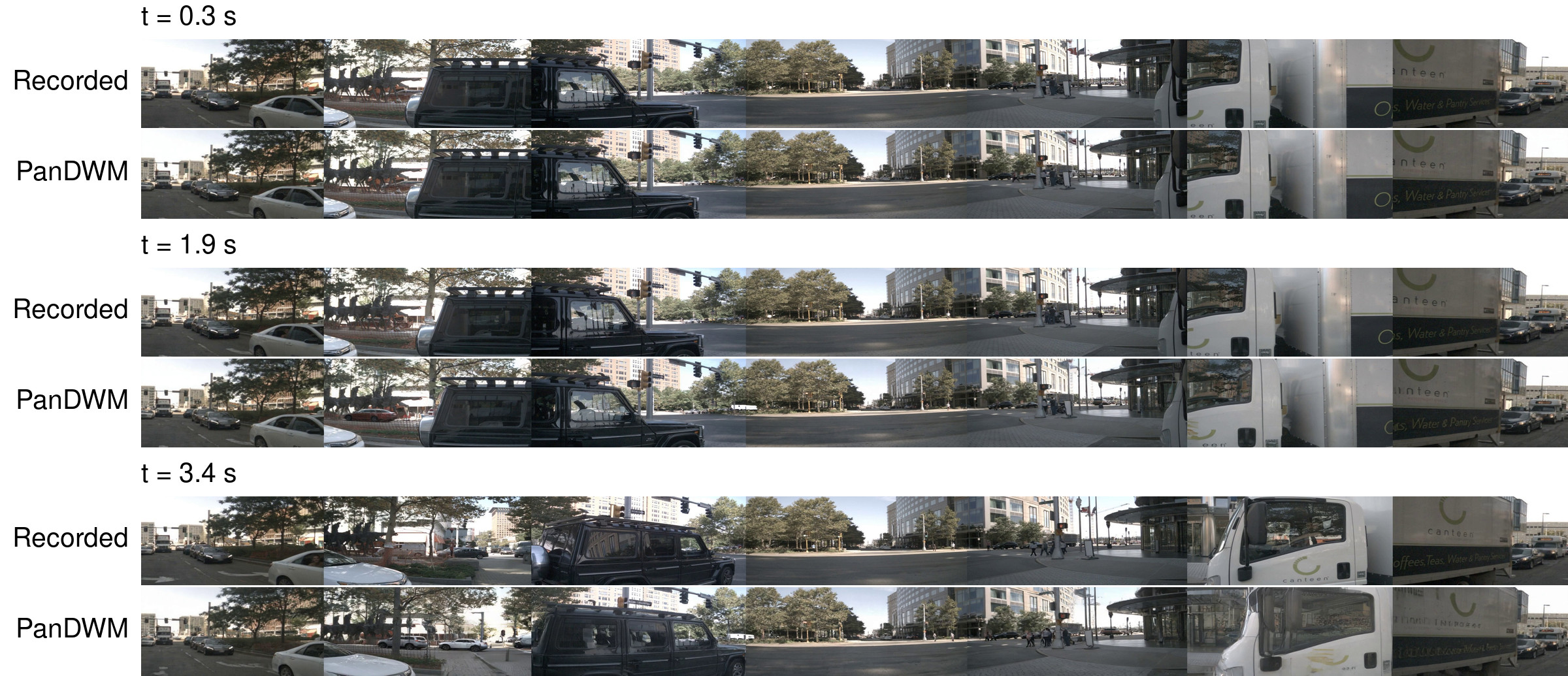}{143.79}{0.00}{Test-S clip 217}
\framepanel{tests_clip217}{71.59}{72.21}{Test-S clip 217}
\framepanel{tests_clip217}{0.00}{144.41}{Test-S clip 217}
\subsection{Test-L rollouts}
\cliptitle{Test-L scene 75, 19.5\,s rollout}
\framepanel{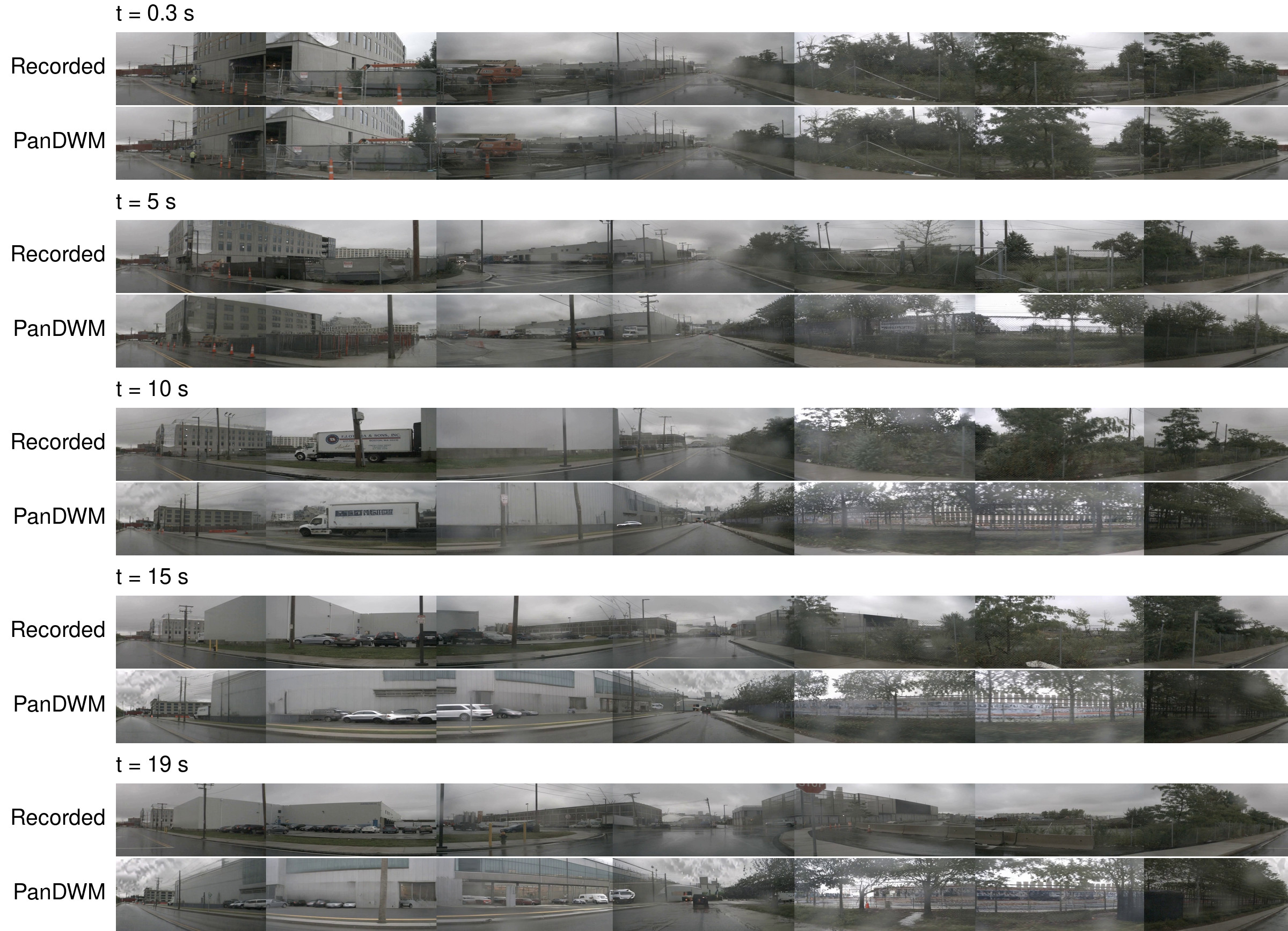}{288.21}{0.00}{Test-L scene 75}
\framepanel{testl_scene075}{216.00}{72.21}{Test-L scene 75}
\framepanel{testl_scene075}{143.79}{144.41}{Test-L scene 75}
\framepanel{testl_scene075}{71.59}{216.62}{Test-L scene 75}
\framepanel{testl_scene075}{0.00}{288.83}{Test-L scene 75}
\cliptitle{Test-L scene 101, 19.5\,s rollout}
\framepanel{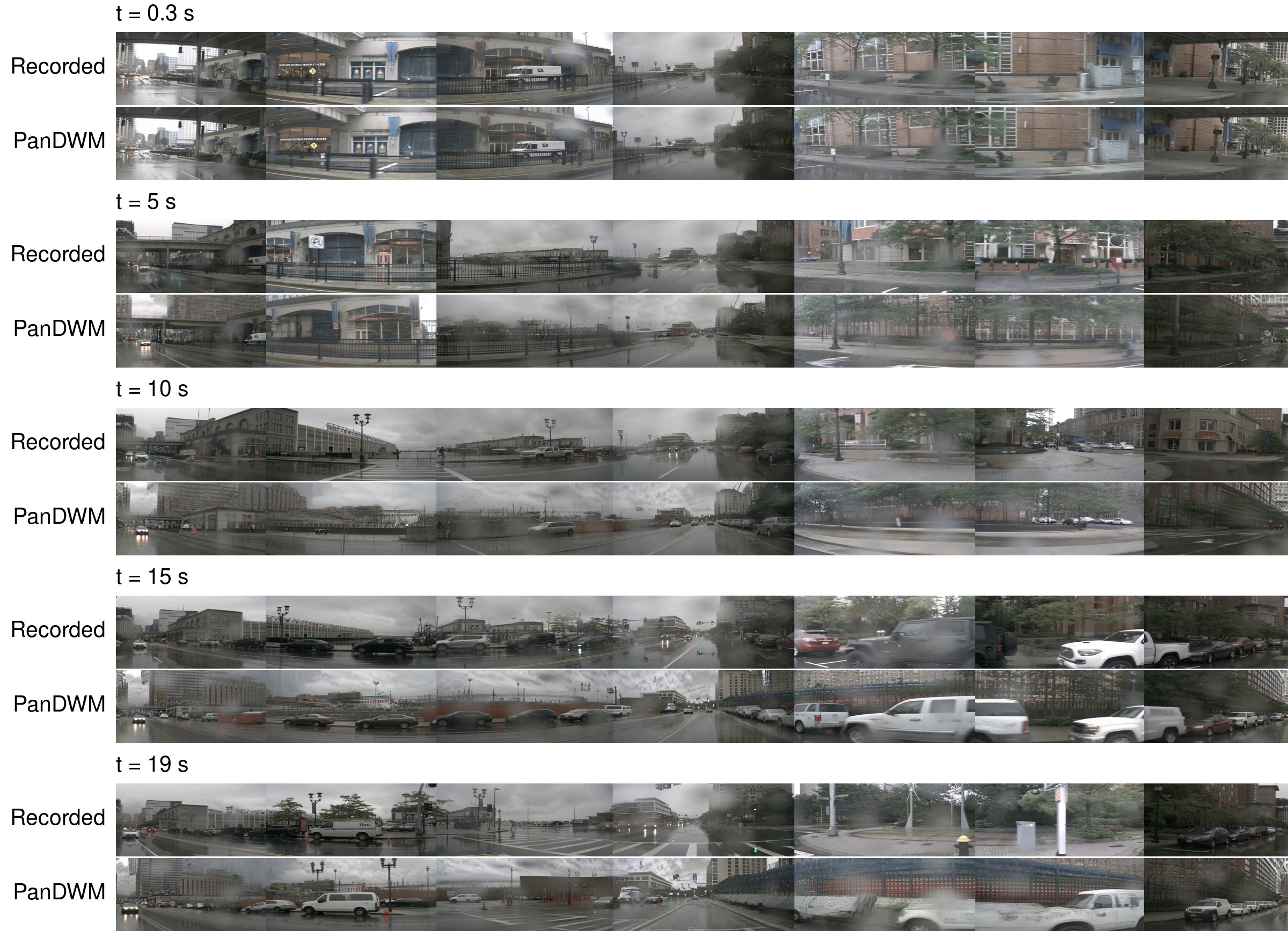}{288.21}{0.00}{Test-L scene 101}
\framepanel{testl_scene101}{216.00}{72.21}{Test-L scene 101}
\framepanel{testl_scene101}{143.79}{144.41}{Test-L scene 101}
\framepanel{testl_scene101}{71.59}{216.62}{Test-L scene 101}
\framepanel{testl_scene101}{0.00}{288.83}{Test-L scene 101}
\cliptitle{Test-L scene 131, 19.5\,s rollout}
\framepanel{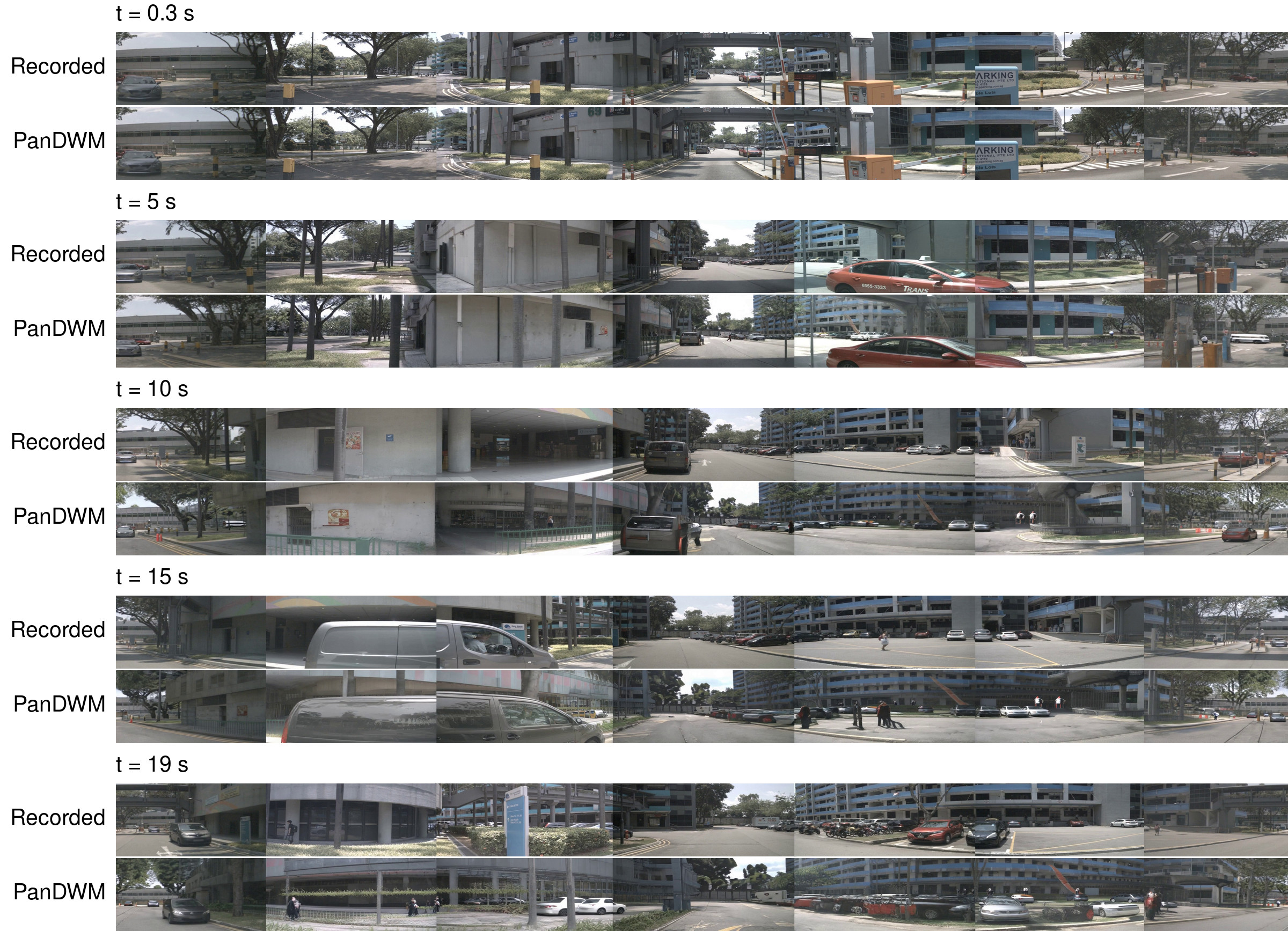}{288.21}{0.00}{Test-L scene 131}
\framepanel{testl_scene131}{216.00}{72.21}{Test-L scene 131}
\framepanel{testl_scene131}{143.79}{144.41}{Test-L scene 131}
\framepanel{testl_scene131}{71.59}{216.62}{Test-L scene 131}
\framepanel{testl_scene131}{0.00}{288.83}{Test-L scene 131}
\cliptitle{Test-L scene 62, 19.5\,s rollout}
\framepanel{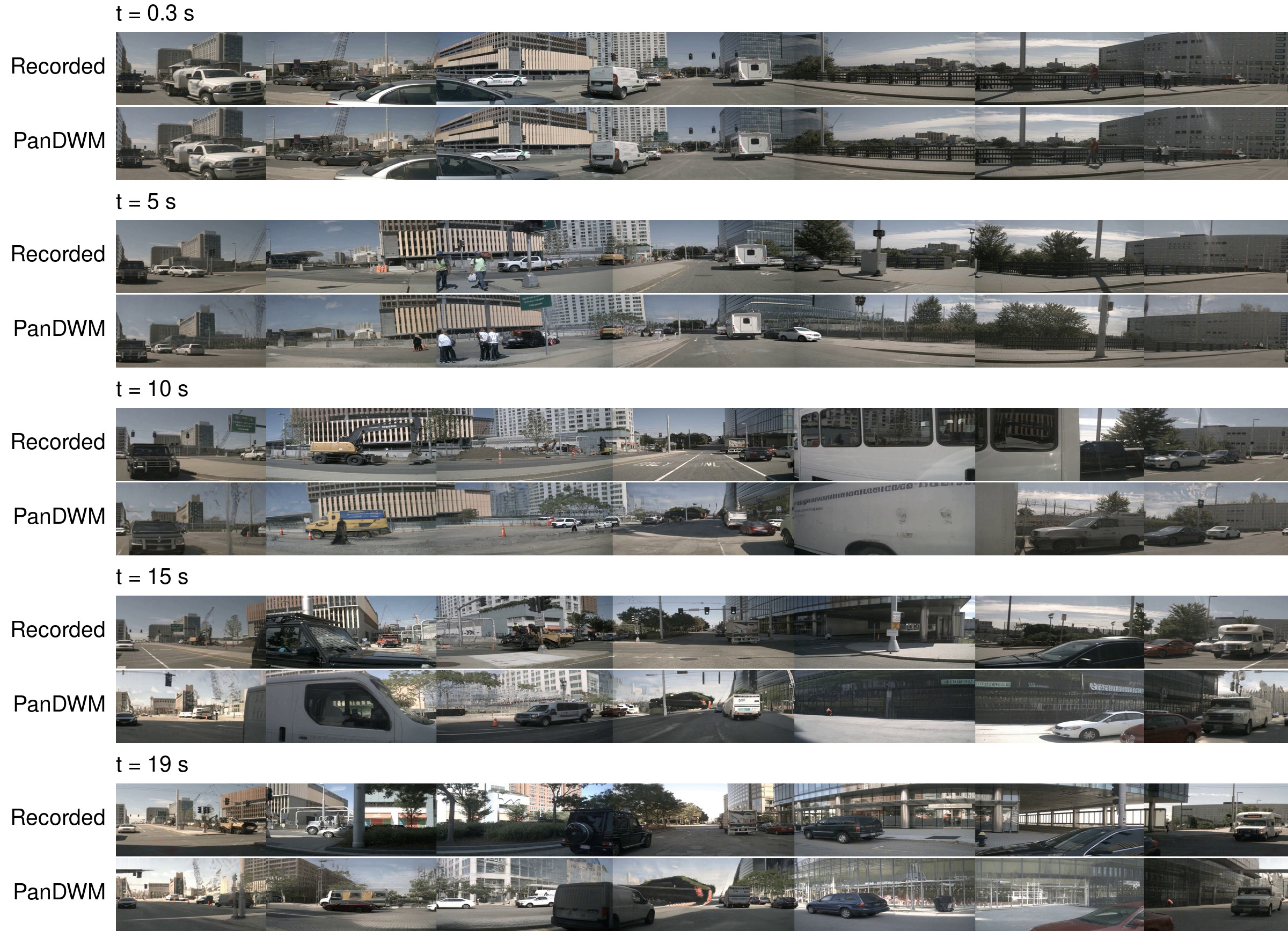}{288.21}{0.00}{Test-L scene 62}
\framepanel{testl_scene062}{216.00}{72.21}{Test-L scene 62}
\framepanel{testl_scene062}{143.79}{144.41}{Test-L scene 62}
\framepanel{testl_scene062}{71.59}{216.62}{Test-L scene 62}
\framepanel{testl_scene062}{0.00}{288.83}{Test-L scene 62}
\cliptitle{Test-L scene 95, 19.5\,s rollout}
\framepanel{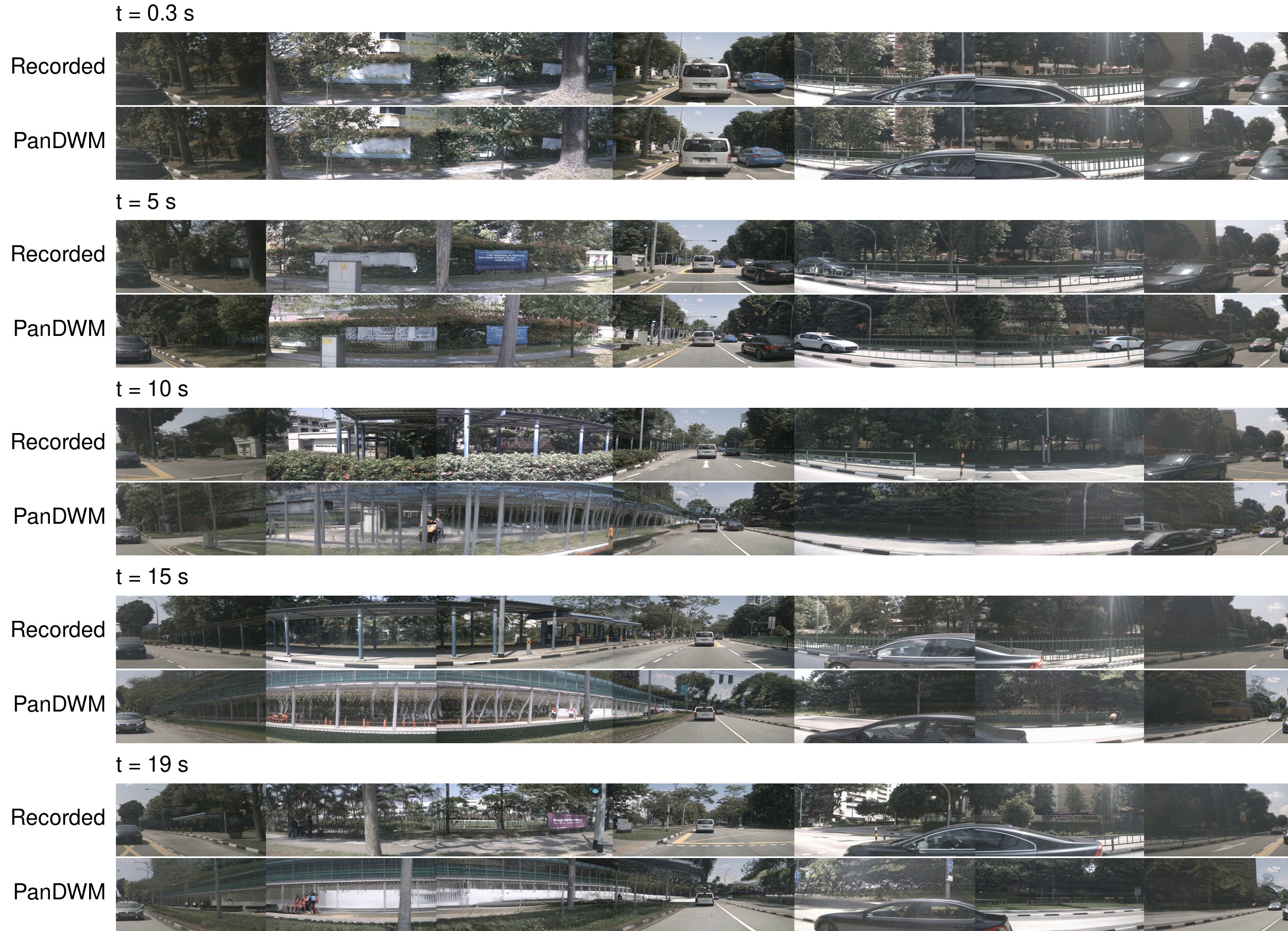}{288.21}{0.00}{Test-L scene 95}
\framepanel{testl_scene095}{216.00}{72.21}{Test-L scene 95}
\framepanel{testl_scene095}{143.79}{144.41}{Test-L scene 95}
\framepanel{testl_scene095}{71.59}{216.62}{Test-L scene 95}
\framepanel{testl_scene095}{0.00}{288.83}{Test-L scene 95}
\cliptitle{Test-L scene 130, 19.5\,s rollout}
\framepanel{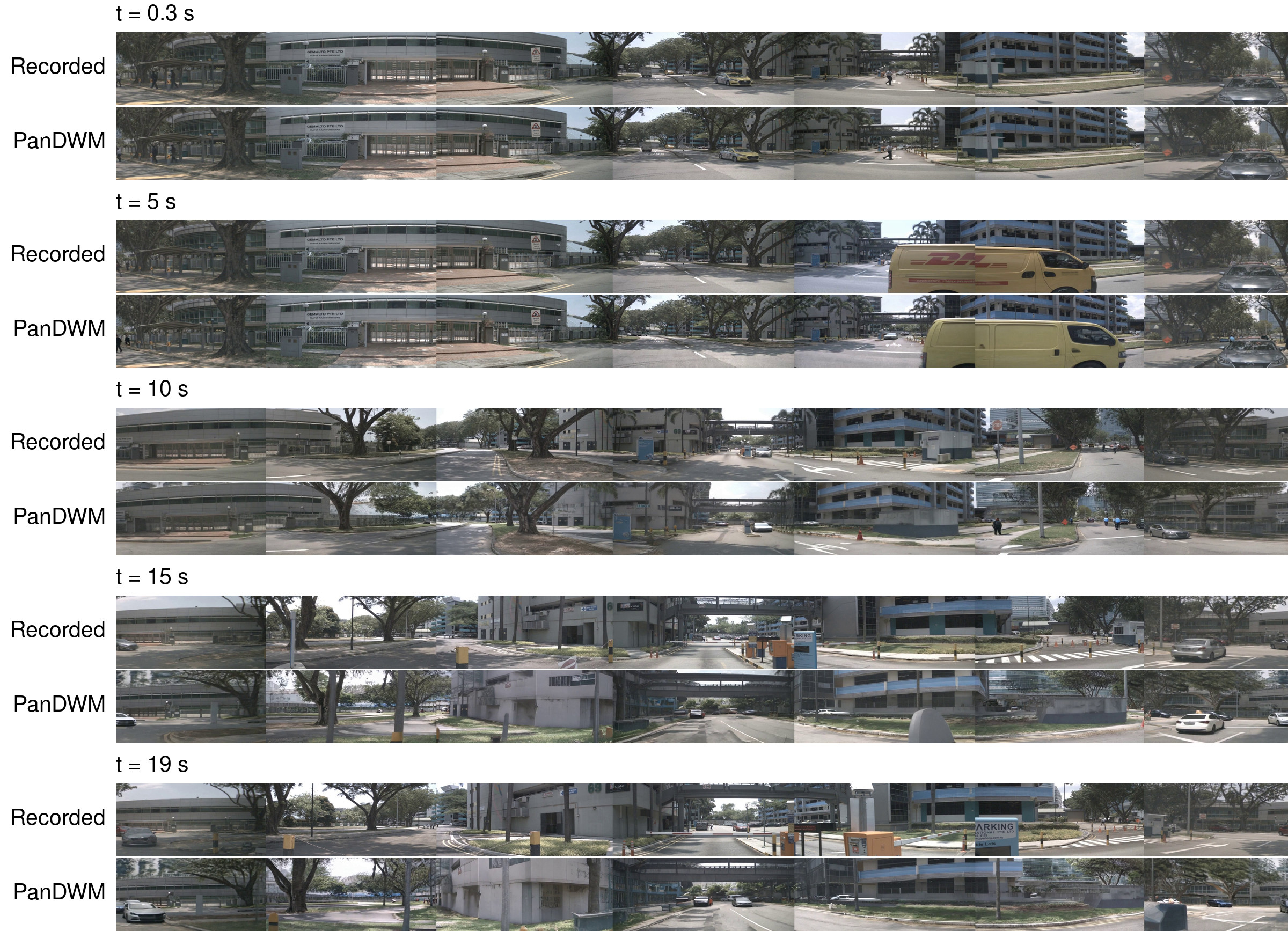}{288.21}{0.00}{Test-L scene 130}
\framepanel{testl_scene130}{216.00}{72.21}{Test-L scene 130}
\framepanel{testl_scene130}{143.79}{144.41}{Test-L scene 130}
\framepanel{testl_scene130}{71.59}{216.62}{Test-L scene 130}
\framepanel{testl_scene130}{0.00}{288.83}{Test-L scene 130}

\clearpage

\end{document}